\pdfoutput=1
\documentclass[11pt]{article}

\usepackage[preprint]{acl}

\usepackage{times}
\usepackage{latexsym}
\usepackage[T1]{fontenc}
\usepackage[utf8]{inputenc}
\usepackage{microtype}
\usepackage{inconsolata}
\usepackage{graphicx}

\usepackage{amsmath,amssymb}
\usepackage{booktabs}
\usepackage{multirow}
\usepackage{makecell}
\usepackage{xcolor}
\definecolor{mapgreen}{HTML}{2CA02C}
\definecolor{mapred}{HTML}{D62728}
\usepackage{colortbl}
\usepackage{pifont}
\usepackage{enumitem}
\usepackage[most]{tcolorbox}

\newcommand{\dataset}{\textsc{GeoDecisions}}
\newcommand{\one}{\ding{182}}
\newcommand{\two}{\ding{183}}
\newcommand{\three}{\ding{184}}
\newcommand{\four}{\ding{185}}

\newcommand{\modelfont}[1]{\texttt{#1}}
\newcommand{\jev}{\modelfont{Jev}}
\newcommand{\mpplx}{\modelfont{pplx-decider-v1.2}}
\newcommand{\mglide}{\modelfont{GLiDE-27B}}
\newcommand{\mknowline}{\modelfont{KnowLine-31B-Gen1}}
\newcommand{\msieve}{\modelfont{Sieve-31B-v3}}
\newcommand{\mjev}{\modelfont{Jev-1.14}}
\newcommand{\mluna}{\modelfont{GPT-6 Luna}}
\newcommand{\mqwenbase}{\modelfont{Qwen3.8-27B}}
\newcommand{\mgemmabase}{\modelfont{Gemma-4-31B-it}}
\newcommand{\mgptfour}{\modelfont{GPT-4 Turbo}}
\newcommand{\mgptthreefive}{\modelfont{GPT-3.5 Turbo}}
\newcommand{\mgeminipro}{\modelfont{Gemini Pro}}
\newcommand{\mmixtral}{\modelfont{Mixtral 8x7B}}
\newcommand{\mllama}{\modelfont{Llama 2 70B}}
\newcommand{\logo}[1]{\raisebox{-0.175em}{\includegraphics[height=0.95em]{logos/#1}}\hspace{0.35em}}

\newcommand{\ssiex}{\ensuremath{\mathrm{SSI}_{\Delta}}}
\newcommand{\sreex}{\ensuremath{\mathrm{SRE}_{\Delta}}}

\setlist{nosep,leftmargin=*}

\title{Seeing Is Locating: Spatial Awareness and\\Geographic Bias in Decision Language Models}

\author{
  Wilson Wongso\textsuperscript{1} \quad
  Hada Melino Muhammad\textsuperscript{1} \quad
  Xiachong Lin\textsuperscript{1} \quad
  Hao Xue\textsuperscript{1,2} \quad
  Flora D. Salim\textsuperscript{1} \\
  \textsuperscript{1}University of New South Wales, Sydney, NSW, Australia \\
  \textsuperscript{2}Hong Kong University of Science and Technology (Guangzhou), Guangzhou, China \\
  \small\texttt{\{w.wongso, flora.salim\}@unsw.edu.au}
}

\begin{document}
\maketitle

\begin{abstract}
    Decision models, which output probability distributions over candidate labels instead of generating text, are increasingly used as classifiers and evaluators for inputs associated with real-world locations. However, it remains unclear whether they can leverage spatial context and whether their predictions are geographically biased.
    We introduce \dataset{}, a multimodal benchmark for evaluating the spatial awareness and geographic bias of decision models. \dataset{} comprises two tasks: \one{} geo-aware image classification, and \two{} location rating on objective and subjective topics.
    Evaluating six decision models against supervised task-specific models, we find that \emph{spatial awareness requires vision}: textual coordinates alone provide insufficient context for classifying everyday photos and satellite images.
    We also find \textit{substantial geographic bias}. On everyday photos, despite achieving higher accuracy, open-weight decision models exhibit more spatially clustered errors than supervised models. Moreover, across all five subjective topics, every decision model gives lower ratings to locations with lower socioeconomic conditions.
    Finally, we propose a training-free method that combines decision models with a location prior to improve their spatial awareness without substantially increasing performance bias.
    Our code is available at \url{https://github.com/cruiseresearchgroup/GeoDecisions}.
\end{abstract}
\section{Introduction}
\label{sec:intro}

\begin{figure}[t]
    \centering
    \includegraphics[width=\columnwidth]{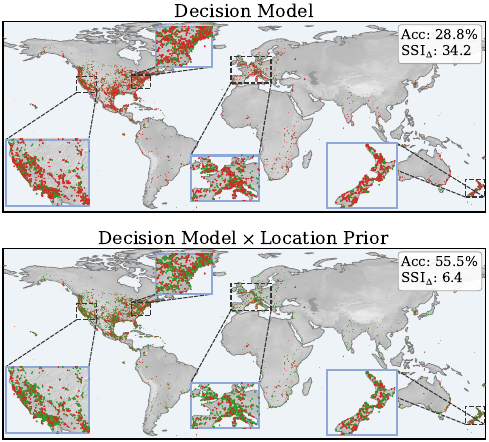}
    \caption{Top-1 predictions of \mpplx{} on iNat2018 in the Image (top) and Image $\times$ Location Prior (bottom) settings (Section~\ref{sec:rq3}). \textcolor{mapgreen}{Green} indicates correct predictions, and \textcolor{mapred}{red} indicates incorrect predictions. Figure design adapted from \citet{wang2025urbanr1}.}
    \label{fig:prior_maps}
    \vspace{-8pt}
\end{figure}

Language models are increasingly used for classification tasks such as labeling, grading, and routing.
\emph{Decision models} such as \jev{} \citep{typesafe2026jev} and its open-weight counterparts \citep{palmer2026kev,perplexity2026decider,sthanika2026sieve} are designed for this use: given an input and a set of candidate labels, they output a probability distribution over the labels instead of generating text, which makes them cheap, fast, and easy to deploy as classifiers and evaluators \citep{rao2026jev,li2026jevjudge}.
Many of these inputs involve real geographic locations, so a \textit{fair} decision model must leverage the spatial context of its input and minimize geographic bias at the same time.

Geographic bias has been studied for generative LLMs \citep{manvi2024large} and for supervised geo-aware models \citep{wu2024torchspatial}, whereas existing studies of decision models focus on how candidate labels are presented \citep{gao2026ordinal,rao2026jev}.
The spatial awareness and geographic bias of decision models therefore remain unexamined, even though they are increasingly used as zero-shot classifiers on geospatial tasks \citep{niu2026geojev,abouelhadid2026eojev}.
We therefore ask three research questions:
\begin{itemize}
    \item \textbf{RQ1: Are decision models spatially aware?} Can they predict from coordinates or place names, and do these inputs complement images?
    \item \textbf{RQ2: Are decision models geographically biased?} Are their errors spatially clustered and do their ratings of people and places correlate with socioeconomic conditions?
    \item \textbf{RQ3: Can decision models be made spatially aware?} Can spatial context be added to decision models without fine-tuning?
\end{itemize}

To answer these questions, we introduce \dataset{}, a multimodal benchmark with two tasks: \one{} \textbf{geo-aware image classification}, adapted from LocBench \citep{wu2024torchspatial}, and \two{} \textbf{location rating}, adapted from \citet{manvi2024large}.
Evaluating six decision models, we find that spatial awareness requires vision (RQ1) and that decision models are geographically biased (RQ2).
We then propose a training-free method that combines a decision model's predictions with a pretrained location prior, which improves spatial awareness (RQ3; Figure~\ref{fig:prior_maps}).
Our contributions are threefold:
\begin{itemize}
    \item \textbf{\dataset{} benchmark.} A multimodal benchmark for decision models and shuffle-corrected geo-bias scores that allow models with different accuracy to be compared.
    \item \textbf{Systematic evaluation of spatial awareness and geographic bias.} An evaluation of open-weight and closed-source decision models, showing that spatial awareness requires vision, that decision models are geographically biased, and that this bias is largely inherited from base LLMs.
    \item \textbf{Training-free spatial awareness.} A model-agnostic method that combines decision models with a location prior to improve spatial awareness without substantially increasing geographic bias.
\end{itemize}

\section{Related Work}
\label{sec:related}

\paragraph{Geographic Bias in Language Models.}
LLMs encode substantial geographic knowledge: the coordinates of named places can be linearly decoded from their representations \citep{gurnee2024language,vafa2024evaluating}, and socioeconomic indicators can be predicted from location prompts \citep{manvi2024geollm}.
Beyond social bias on attributes such as gender and race \citep{dhamala2021bold,parrish2022bbq}, geography offers a view of bias with measurable ground truth \citep{manvi2024large}.
LLMs recall geographic knowledge unevenly across countries \citep{moayeri2024worldbench}, over-amplify geopolitical favoritism \citep{faisal2023geographic}, and rate locations with lower socioeconomic conditions lower on sensitive subjective topics \citep{manvi2024large}. Likewise, vision and vision-language models (VLMs) are less accurate on images from lower-income regions \citep{devries2019does,shankar2017no,nwatu2023bridging}.
Urban-R1 \citep{wang2025urbanr1} mitigates such bias by fine-tuning VLMs with reinforcement learning via GRPO \citep{deepseekai2024deepseek}.
These studies target generative models, and mitigating bias through fine-tuning is computationally costly and requires access to model weights, which closed-source decision models such as \jev{} do not provide.
We instead evaluate decision models and improve their spatial awareness with a training-free method that applies to both open-weight and closed-source models (Section~\ref{sec:method}).

\paragraph{Geo-Aware Benchmarks.}
Location encoders improve geo-aware tasks such as species recognition and remote sensing image classification \citep{macaodha2019presence,mai2020space2vec,mai2023sphere2vec,mai2022review} on fMoW \citep{christie2018fmow}, iNaturalist \citep{vanhorn2018inaturalist}, and YFCC \citep{tang2015improving}.
TorchSpatial \citep{wu2024torchspatial} unifies 15 location encoders in the LocBench benchmark and introduces geo-bias scores based on spatial self-information \citep{wang2024probing}, later extended with spatial relative entropy in GeoBS \citep{wang2025geobs}.
These benchmarks show that location encoders increase geographic bias alongside accuracy, but evaluate only supervised models trained for each task.
To our knowledge, no existing benchmark evaluates both the spatial awareness and geographic bias of (zero-shot) decision models, a gap that \dataset{} fills by providing a unified evaluation framework decision models.

\paragraph{Decision Models.}
Decision (language) models output a probability distribution over a set of candidate labels instead of generating an answer or scoring answer tokens \citep{holtzman2021surface,robinson2023leveraging}.
\jev{} \citep{typesafe2026jev} offers this as a closed-source API, and several open-weight models adapt Qwen and Gemma LLMs into decision models through fine-tuning \citep{palmer2026kev,perplexity2026decider,fastino2026glide,pelaai2026knowline,sthanika2026sieve}.
Existing studies compare \jev{} with LLM judges \citep{rao2026jev,li2026jevjudge} and show that decision models underuse ordinal rating scales \citep{gao2026ordinal}, echoing the label and position biases of LLMs \citep{zhao2021calibrate,zheng2024large,pezeshkpour2024large,zheng2023judging,stureborg2024large}.
However, these studies only examine how labels are presented. Although decision models are increasingly used on geospatial tasks \citep{niu2026geojev,abouelhadid2026eojev}, whether they are spatially aware or geographically biased remains understudied.

\section{Problem Formulation}
\label{sec:prelim}

\paragraph{Decision Models.}
Each sample consists of an input $x=(q, l, I)$ (a question $q$, an optional location $l$, e.g., coordinates or a place name, and an optional image $I$) and a set of $K$ candidate labels $\mathcal{Y}=\{y_1,\ldots,y_K\}$.
A decision model $f_\theta$ assigns each candidate label $y\in\mathcal{Y}$ a probability $f_\theta(y \mid x, \mathcal{Y})$, where $\sum_{y\in\mathcal{Y}} f_\theta(y \mid x, \mathcal{Y})=1$, and predicts the most probable label,
\begin{equation}
    \hat{y}=\arg\max_{y\in\mathcal{Y}} f_\theta(y \mid x, \mathcal{Y}).
\end{equation}

We consider two sample types: \emph{classification}, where the labels are unordered classes, and \emph{rating}, where the labels are ordered numeric values.
For rating samples, we instead take the expected value of the rating, $\hat{v}=\sum_{y\in\mathcal{Y}} y\, f_\theta(y \mid x, \mathcal{Y})$ as the prediction, following \citet{manvi2024large}.

Unlike open-weight models, closed-source models can refuse a sample and return no prediction.
We count a refused classification sample as an error and exclude refused rating samples from the predictions, accounting for them through the answer rate (see Section~\ref{sec:metrics}).

\paragraph{Spatial Awareness and Geographic Bias.}
A decision model is \emph{spatially aware} if it leverages spatial context: it predicts correctly from the location $l$ alone, and adding $l$ improves its predictions given the image $I$.
Following \citet{wu2024torchspatial} and \citet{manvi2024large}, we study two forms of \emph{geographic bias}: \one{} \emph{performance bias}, where model errors are disproportionately clustered in certain geographic regions, and \two{} \emph{social bias}, where ratings of subjective topics correlate with regional socioeconomic conditions despite being location-independent.
\section{\dataset{} Benchmark}
\label{sec:benchmark}

\begin{figure*}[!t]
    \centering
    \scriptsize
    \tcbset{exbox/.style={colback=gray!4,colframe=gray!60,boxrule=0.4pt,left=3pt,right=3pt,lefttitle=2pt,righttitle=1pt,top=2pt,bottom=2pt,
                fonttitle=\scriptsize,colbacktitle=gray!25,coltitle=black,equal height group=examples,
                before upper={\raggedright}}}
    \begin{minipage}[t]{0.25\textwidth}
        \begin{tcolorbox}[exbox,title={\textbf{Task 1}: Coordinates}]
            \textbf{Input:} \texttt{Location: latitude 39.5425, longitude -77.1073} \\
            \textbf{Question:} Which land use or building category is most likely to be found at this location? \\
            \textbf{Labels:} airport, airport hangar, \ldots, zoo \\
            \textbf{Ground truth:} burial site
        \end{tcolorbox}
    \end{minipage}\hfill
    \begin{minipage}[t]{0.25\textwidth}
        \begin{tcolorbox}[exbox,title={\mbox{\textbf{Task 1}: Image + Coordinates}}]
            \textbf{Input:}\par\vspace{1pt}\noindent\includegraphics[width=\linewidth]{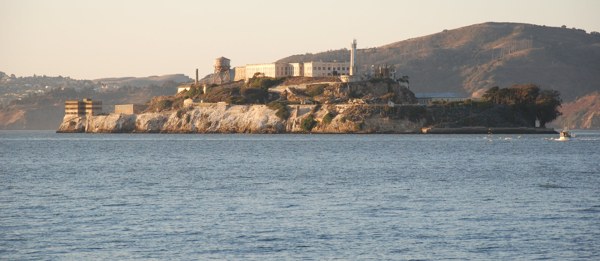}\par{\centering\tiny\itshape ``Look, its Alcatraz'' by Mike Renlund, \href{https://www.flickr.com/photo.gne?id=1449342277}{Flickr}, \href{https://creativecommons.org/licenses/by/2.0/}{CC BY 2.0}\par}\vspace{1pt}+ \texttt{Location: latitude 37.8101, longitude -122.4171} \\
            \textbf{Question:} Which of these concepts best describes this photo? \\
            \textbf{Labels:} airplane, \ldots, yacht \\
            \textbf{Ground truth:} alcatraz
        \end{tcolorbox}
    \end{minipage}\hfill
    \begin{minipage}[t]{0.243\textwidth}
        \begin{tcolorbox}[exbox,title={\textbf{Task 2}: Coordinates}]
            \textbf{Input:} \texttt{Coordinates: (40.77042, -73.95542)} \\
            \textbf{Question:} You will be given data about a specific location randomly sampled from all human-populated locations on Earth. You give your rating keeping in mind that it is relative to all other human-populated locations on Earth (from all continents, countries, etc.). Rate: Average Intelligence of Residents (on a scale from 0.0 to 9.9). \\
            \textbf{Labels:} 0.0, 0.1, \ldots, 9.9 \\
            \textbf{Ground truth:} none \textit{(subjective topic)}
        \end{tcolorbox}
    \end{minipage}\hfill
    \begin{minipage}[t]{0.243\textwidth}
        \begin{tcolorbox}[exbox,title={\textbf{Task 2}: Full GeoLLM Prompt}]
            \textbf{Input:} \texttt{Coordinates: (3.46208, 35.91958)} \newline \texttt{Address: "Turkana, Kenya"} \newline \texttt{Nearby Places: 10.5 km North-West: Kalokol; \ldots} \\
            \textbf{Question:} You will be given data about [\ldots{}] Rate: Average Temperature (on a scale from 0.0 to 9.9). \\
            \textbf{Labels:} 0.0, 0.1, \ldots, 9.9 \\
            \textbf{Ground truth:} 29.9\,$^\circ$C
        \end{tcolorbox}
    \end{minipage}
    \caption{Example samples from \dataset{} illustrating the inputs, questions, and candidate labels for both tasks. Task 1 examples show two input configurations: coordinates only and image with coordinates. Task 2 examples show a coordinates-only input and a full GeoLLM prompt.}
    \label{fig:examples}
\end{figure*}

In this section, we introduce the two benchmark tasks in \dataset{} and describe their corresponding evaluation metrics.
\textbf{Task 1: geo-aware image classification} evaluates a model's ability to classify an image given the image itself, its geographic coordinates, or both, examining whether decision models exhibit spatial awareness (RQ1).
\textbf{Task 2: location rating} evaluates a model's ability to rate a location on both objective and subjective topics given its coordinates, place name, or both, examining whether decision models exhibit geographic bias (RQ2).
On both tasks, we evaluate spatial awareness using accuracy metrics and geographic bias using geo-bias metrics, which assess whether model errors and ratings exhibit spatial clustering or correlations with socioeconomic conditions.
Table~\ref{tab:stats} summarizes the dataset statistics and Figure~\ref{fig:examples} shows examples of both tasks.

\begin{table}[t]
    \centering
    \small
    \setlength{\tabcolsep}{3pt}
    \begin{tabular}{@{}llrrr@{}}
        \toprule
        \textbf{Dataset}                                         & \textbf{Domain}  & \textbf{\#Samples} & \textbf{\#Classes} & $K$ \\
        \midrule
        \multicolumn{5}{@{}l}{\textit{Task 1: geo-aware image classification}}                                                      \\
        fMoW                                                     & satellite images & 53,041             & 62                 & 62  \\
        YFCC                                                     & everyday photos  & 17,798             & 100                & 100 \\
        iNat2017                                                 & species photos   & 95,986             & 5,089              & 20  \\
        iNat2018                                                 & species photos   & 24,426             & 8,142              & 20  \\
        \midrule
        \multicolumn{5}{@{}l}{\textit{Task 2: location rating}}                                                                     \\
        Objective                                                & 6 topics         & 2,000              & --                 & 100 \\
        Subjective                                               & 5 topics         & 2,000              & --                 & 100 \\
        \midrule
        \multicolumn{4}{@{}l}{Task 1 (3 input settings)}         & 568,857                                                          \\
        \multicolumn{4}{@{}l}{Task 2 (3 input settings)}         & 66,000                                                           \\
        \multicolumn{4}{@{}l}{\textbf{Total evaluation samples}} & \textbf{634,857}                                                 \\
        \bottomrule
    \end{tabular}
    \caption{Dataset statistics for \dataset{}. \#Samples denotes the number of images in Task 1 or locations in Task 2. \#Classes is the number of classification classes, while $K$ denotes the number of candidate classes (Task 1) or rating levels (Task 2).}
    \label{tab:stats}
\end{table}

\paragraph{Data Contamination.}
As both source benchmarks are public, we run three training-data contamination tests \citep{song2024mmdetect,golchin2023quiz,seshadri2022quantifying} and find minimal evidence of memorization or template tuning, apart from a small original-order gain for \mluna{} (see Appendix~\ref{app:contamination} for details).
\subsection{Task 1: Geo-Aware Image Classification}
\label{sec:track1}

\paragraph{Datasets.}
We adopt the four image classification datasets of LocBench \citep{wu2024torchspatial}, which span different image domains and geographic coverages: \one{} fMoW \citep{christie2018fmow}, a global remote sensing dataset of satellite images labeled with 62 land-use and building categories; \two{} YFCC \citep{tang2015improving}, a dataset of geotagged photos labeled with 100 everyday concepts, mostly taken in the United States; and \three{} iNat2017 and \four{} iNat2018 \citep{vanhorn2018inaturalist}, two species recognition datasets with thousands of fine-grained species observed across all continents.
We use the same evaluation splits as LocBench/TorchSpatial.

\paragraph{Input Settings and Candidate Labels.}
Each sample is evaluated in three input settings: \one{} \emph{Image}, where the model sees only the image; \two{} \emph{Coordinates}, where the model sees only the coordinates; and \three{} \emph{Image + Coordinates}, where the model sees both.
In the two image settings, the model is asked what the image shows, whereas in the Coordinates setting, it is asked which class is most likely found at the location (Appendix~\ref{app:prompts}).
For fMoW and YFCC, every class is a candidate label.
The iNat2017 and iNat2018 datasets contain 5,089 and 8,142 classes, respectively, exceeding the candidate-label capacity of most decision models. We therefore restrict each sample's candidate labels to the top-20 classes predicted by the TorchSpatial image-only model, which include the ground-truth class for 92\% of samples. Candidate labels are shuffled per sample using a fixed seed.

\subsection{Task 2: Location Rating}
\label{sec:track2}

\paragraph{Datasets.}
We adopt the 2,000 world locations released by \citet{manvi2024large}, which were sampled using their population-weighted importance sampling and farthest point sampling.
Each location is rated on 6 \emph{objective} topics (average temperature, annual precipitation, infant mortality rate, population density, nighttime light intensity, built-up area ratio), whose ground truth is aggregated from global rasters over a 25$\times$25-pixel window around the location, following GeoLLM (Appendix~\ref{app:groundtruth}).
Each location is also rated on 5 \emph{subjective} topics without ground truth (average attractiveness, intelligence, likability, morality, work ethic of residents), on which an unbiased model gives a constant or random rating at every location \citep{manvi2024large}.

\paragraph{Input Settings and Candidate Labels.}
Each location is evaluated under three input settings: \one{} \emph{Coordinates}, where the model receives only the geographic coordinates; \two{} \emph{Place Name}, where it receives only the address; and \three{} \emph{Full}, where it receives the complete GeoLLM prompt, including coordinates, address, and nearby places \citep{manvi2024geollm}.
In all three settings, the model is asked to rate the location on a given topic using the prompt from \citet{manvi2024large} (Figure~\ref{fig:examples}; Appendix~\ref{app:prompts}).
The candidate labels are the 100 ratings from 0.0 to 9.9, and the prediction is the expected value of the rating under the model's probability distribution over these labels (see Section~\ref{sec:prelim}).

\subsection{Evaluation Metrics}
\label{sec:metrics}

\paragraph{Accuracy.}
For Task 1, we report top-1 accuracy.
For objective topics in Task 2, we report the Spearman rank correlation $\rho$ between predicted ratings and ground truth \citep{manvi2024large}.

\paragraph{Shuffle-Corrected Geo-Bias Scores.}
To measure performance bias, we assign each sample $i$ a performance value $m_i$ (its \emph{mark}) at its location $\ell_i$ and test whether marks are spatially structured around a set of evaluation centers, using the geo-bias scores of \citet{wu2024torchspatial} and \citet{wang2025geobs}.
Briefly, \emph{SSI} (spatial self-information) measures how unlikely the spatial autocorrelation of marks within each center's neighborhood is under a random arrangement, and \emph{SRE} (spatial relative entropy) divides each neighborhood into square cells and averages the KL divergence between the mark distribution of each cell and that of the neighborhood.
Both scores, however, also depend on the overall mark distribution: a model that is wrong almost everywhere obtains a high SSI regardless of where its errors fall.
We therefore subtract the score obtained after randomly permuting the marks across samples, which preserves their distribution but removes their spatial structure:
\begin{equation}
    \mathrm{S}_{\Delta} = \mathrm{S}(\{(\ell_i,m_i)\}_{i=1}^N) - \mathrm{S}(\{(\ell_i,m_{\pi(i)})\}_{i=1}^N),
\end{equation}
where $\mathrm{S}$ is SSI or SRE and $\pi$ is a fixed random permutation.
\ssiex{} and \sreex{} are zero in expectation when errors are spatially random, regardless of accuracy, and higher values indicate stronger clustering.
We compute \textit{performance bias} scores for Task~1 and use Hit@1 as the mark.
Other evaluation metric parameters are provided in Appendix~\ref{app:excess}.

\paragraph{Social Bias Score.}
For subjective topics, we use the bias score of \citet{manvi2024large}:
\begin{equation}
    B(\hat{\boldsymbol{v}}) = \rho(\hat{\boldsymbol{v}},\boldsymbol{a})\cdot\mathrm{MAD}(\hat{\boldsymbol{v}})\cdot \alpha^2,
\end{equation}
where $\hat{\boldsymbol{v}}$ are the predicted ratings, $\boldsymbol{a}$ is the infant mortality rate at each location as a proxy for socioeconomic conditions, MAD is the mean absolute deviation, and $\alpha$ is the answer rate.
A negative $B$ therefore means that a model gives lower ratings in regions with lower socioeconomic conditions.

\section{Training-Free Location Priors for Decision Models}
\label{sec:method}

\begin{table*}[!t]
    \centering
    \small
    \setlength{\tabcolsep}{3.0pt}
    \begin{tabular}{@{}l cccc|cccc|cccc|cccc@{}}
        \toprule
                                         & \multicolumn{4}{c|}{\textbf{YFCC}} & \multicolumn{4}{c|}{\textbf{fMoW}} & \multicolumn{4}{c|}{\textbf{iNat2017}} & \multicolumn{4}{c@{}}{\textbf{iNat2018}}                                                                                                                                                                                                                \\
        \cmidrule(lr){2-5}\cmidrule(lr){6-9}\cmidrule(lr){10-13}\cmidrule(l){14-17}
        \textbf{Model}                   & \textbf{I}                         & \textbf{C}                         & \textbf{I+C}                           & \textbf{I$\times$P}                      & \textbf{I}       & \textbf{C} & \textbf{I+C} & \textbf{I$\times$P} & \textbf{I}       & \textbf{C} & \textbf{I+C} & \textbf{I$\times$P} & \textbf{I}       & \textbf{C} & \textbf{I+C} & \textbf{I$\times$P} \\
        \midrule
        TorchSpatial                     & 50.1                               & 12.1                               & N/A$^*$                                & 51.2                                     & \underline{69.8} & 17.2       & N/A$^*$      & \textbf{70.5}       & \underline{63.3} & 30.4       & N/A$^*$      & \textbf{68.3}       & \underline{60.2} & 48.9       & N/A$^*$      & \textbf{73.5}       \\
        \midrule
        \logo{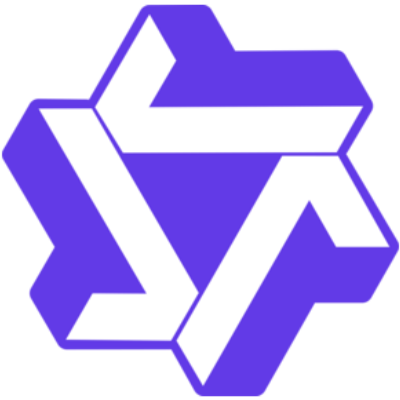}\mqwenbase{}          & 62.0                               & 5.4                                & 62.6                                   & 61.9                                       & 35.1             & 2.2        & 35.9         & 39.5                  & 37.6             & 13.7       & 42.0         & 52.3                  & 25.8             & 13.1       & 31.1         & 52.8                  \\
        \logo{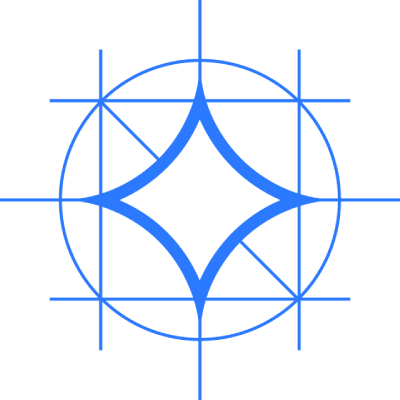}\mgemmabase{}        & 59.9                               & 3.8                                & 60.6                                   & 60.0                                       & 39.9             & 3.9        & 40.9         & 40.7                  & 36.2             & 12.5       & 41.9         & 45.1                  & 24.8             & 13.1       & 30.6         & 42.4                  \\
        \midrule
        \logo{qwen}\mpplx{}              & 63.4                               & 6.4                                & 63.3                                   & 63.1                                     & 38.3             & 2.6        & 38.7         & 43.2                & 40.3             & 17.4       & 46.4         & 54.3                & 28.8             & 17.0       & 35.5         & 55.5                \\
        \logo{qwen}\mglide{}             & 63.6                               & 6.8                                & 63.8                                   & 63.4                                     & 35.4             & 3.4        & 35.8         & 39.2                & 40.1             & 19.0       & 46.1         & 53.9                & 28.4             & 18.5       & 35.4         & 54.8                \\
        \logo{gemma}\mknowline{}         & 62.8                               & 3.8                                & 62.8                                   & 62.1                                     & 42.4             & 4.4        & 43.6         & 48.4                & 37.9             & 14.0       & 43.5         & 51.3                & 26.0             & 14.6       & 32.1         & 51.9                \\
        \logo{gemma}\msieve{}            & \underline{64.4}                   & 4.4                                & \textbf{64.8}                          & \underline{64.4}                         & 42.7             & 2.7        & 43.3         & 48.2                & 40.7             & 11.7       & 46.1         & 53.6                & 28.5             & 13.0       & 34.3         & 53.7                \\
        \midrule
        \logo{openai}\mluna{}$^\dagger$  & 55.6                               & 5.9                                & 55.0                                   & 56.0                                     & 30.5             & 6.2        & 32.2         & 35.5                & 43.8             & 21.8       & 52.3         & 56.5                & 33.0             & 23.3       & 45.6         & 57.5                \\
        \logo{typesafe}\mjev{}$^\dagger$ & 62.0                               & 3.9                                & 61.6                                   & 61.8                                     & 34.6             & 4.6        & 35.8         & 40.2                & 43.9             & 15.7       & 49.2         & 57.1                & 31.7             & 16.6       & 38.4         & 57.2                \\
        \bottomrule
    \end{tabular}
    \caption{Top-1 accuracy (\%, $\uparrow$) on Task 1 in the Image (I), Coordinates (C), Image + Coordinates (I+C), and Image $\times$ Location Prior (I$\times$P) settings. Bold and underlined values indicate the \textbf{best} and \underline{second-best} results per dataset, respectively. $^*$TorchSpatial models do not take coordinates as textual input. $^\dagger$Evaluated on a 2,000-sample subset.}
    \label{tab:track1_acc}
\end{table*}

\begin{table*}[!t]
\centering
\small
\setlength{\tabcolsep}{2.6pt}
\resizebox{\textwidth}{!}{%
\begin{tabular}{@{}l >{\centering\arraybackslash}p{2.3em}>{\centering\arraybackslash}p{2.3em}>{\centering\arraybackslash}p{2.3em}|>{\centering\arraybackslash}p{2.3em}>{\centering\arraybackslash}p{2.3em}>{\centering\arraybackslash}p{2.3em}|>{\centering\arraybackslash}p{2.3em}>{\centering\arraybackslash}p{2.3em}>{\centering\arraybackslash}p{2.3em}|>{\centering\arraybackslash}p{2.3em}>{\centering\arraybackslash}p{2.3em}>{\centering\arraybackslash}p{2.3em}|>{\centering\arraybackslash}p{2.3em}>{\centering\arraybackslash}p{2.3em}>{\centering\arraybackslash}p{2.3em}|>{\centering\arraybackslash}p{2.3em}>{\centering\arraybackslash}p{2.3em}>{\centering\arraybackslash}p{2.3em}@{}}
\toprule
 & \multicolumn{3}{c|}{\textbf{Temperature}} & \multicolumn{3}{c|}{\textbf{Infant Mortality}} & \multicolumn{3}{c|}{\textbf{Precipitation}} & \multicolumn{3}{c|}{\textbf{Nighttime Lights}} & \multicolumn{3}{c|}{\textbf{Population}} & \multicolumn{3}{c@{}}{\textbf{Built-up}} \\
\cmidrule(lr){2-4}\cmidrule(lr){5-7}\cmidrule(lr){8-10}\cmidrule(lr){11-13}\cmidrule(lr){14-16}\cmidrule(lr){17-19}
\textbf{Model} & \textbf{C} & \textbf{P} & \textbf{F} & \textbf{C} & \textbf{P} & \textbf{F} & \textbf{C} & \textbf{P} & \textbf{F} & \textbf{C} & \textbf{P} & \textbf{F} & \textbf{C} & \textbf{P} & \textbf{F} & \textbf{C} & \textbf{P} & \textbf{F} \\
\midrule
\logo{qwen}\mqwenbase{} & 0.81 & 0.92 & 0.94 & 0.70 & 0.91 & 0.91 & 0.59 & 0.91 & 0.91 & 0.23 & 0.77 & 0.78 & 0.35 & 0.77 & 0.79 & 0.30 & 0.76 & 0.74 \\
\logo{gemma}\mgemmabase{} & 0.85 & 0.93 & 0.94 & 0.78 & 0.88 & 0.88 & 0.63 & 0.92 & \underline{0.93} & 0.17 & 0.81 & 0.82 & 0.31 & 0.76 & 0.80 & 0.25 & 0.72 & \underline{0.77} \\
\midrule
\logo{qwen}\mpplx{} & 0.88 & 0.93 & 0.94 & 0.84 & \textbf{0.93} & \textbf{0.93} & 0.73 & 0.91 & 0.91 & 0.28 & 0.80 & 0.82 & 0.35 & 0.78 & \underline{0.81} & 0.28 & 0.73 & 0.75 \\
\logo{qwen}\mglide{} & 0.86 & 0.94 & 0.94 & 0.83 & \underline{0.92} & \underline{0.92} & 0.70 & 0.92 & 0.92 & 0.30 & 0.83 & 0.85 & 0.34 & 0.80 & \textbf{0.83} & 0.24 & \underline{0.77} & \textbf{0.78} \\
\logo{gemma}\mknowline{} & 0.87 & 0.92 & 0.90 & 0.83 & 0.90 & 0.88 & 0.73 & \textbf{0.94} & \textbf{0.94} & 0.32 & 0.81 & 0.82 & 0.34 & 0.74 & 0.72 & 0.36 & 0.71 & 0.72 \\
\logo{gemma}\msieve{} & 0.85 & \underline{0.96} & \textbf{0.97} & 0.81 & 0.90 & 0.89 & 0.70 & \textbf{0.94} & \textbf{0.94} & 0.23 & 0.79 & 0.80 & 0.34 & 0.76 & 0.77 & 0.30 & 0.69 & 0.74 \\
\midrule
\logo{openai}\mluna{} & 0.85 & 0.91 & 0.91 & 0.71 & 0.59 & 0.71 & 0.70 & 0.87 & 0.89 & 0.70 & 0.83 & 0.85 & 0.70 & 0.80 & \textbf{0.83} & 0.67 & 0.76 & \underline{0.77} \\
\logo{typesafe}\mjev{} & 0.82 & 0.92 & 0.92 & 0.69 & 0.90 & 0.90 & 0.75 & 0.92 & \underline{0.93} & 0.31 & \underline{0.87} & \textbf{0.88} & 0.40 & 0.77 & \underline{0.81} & 0.36 & 0.75 & 0.76 \\
\bottomrule
\end{tabular}}
\caption{Spearman $\rho$ ($\uparrow$) on the objective topics of Task 2 (temperature, infant mortality, precipitation, nighttime lights, population density, and built-up ratio) in the Coordinates (C), Place Name (P), and Full (F) settings. Bold and underlined values indicate the \textbf{best} and \underline{second-best} results per topic, respectively.}
\label{tab:track2_obj}
\end{table*}

Despite encoding rich geographic knowledge about named places, multimodal language models struggle to leverage geographic coordinates as textual inputs, a limitation also observed in decision models (Section~\ref{sec:rq1}).
To address this limitation, we propose a \textbf{training-free} method that combines a decision model's predictions with a location prior from a pretrained location encoder.
Unlike generative LLMs, which produce free-form text, decision models output probability distributions over candidate labels, enabling direct probabilistic composition with an external location prior. We exploit this property to combine a zero-shot decision model with a pretrained location encoder, treating image and location as conditionally independent sources of evidence for the label.

Formally, given a sample with image $I$, location $l$, and $K$ candidate labels $\mathcal{Y}$, we query the decision model using only the image and reweight its predictions with a location prior $g$:
\begin{equation}
    p(y \mid I, l) \propto f_\theta(y \mid I, \mathcal{Y}) \cdot g(y \mid l),
    \label{eq:locprior}
\end{equation}
where $g(y\mid l)$ denotes the probability of observing label $y$ at location $l$, as estimated by the location encoder, and the product is normalized over $y \in \mathcal{Y}$. $g$ is evaluated only over the candidate labels and supports an arbitrary number of candidate labels $K$. It requires no parameter updates and uses only the decision model's output probabilities, making it applicable to open-weight and closed-source models.

Our method therefore represents location as a distribution over candidate labels rather than as text prompts.
Unlike existing supervised geo-aware models \citep{macaodha2019presence,wu2024torchspatial}, our method uses a zero-shot decision model without training on the target dataset. We apply this method to Task~1, where location provides spatial context for image classification, as a fourth input setting \four{} \emph{Image $\times$ Location Prior}.
We do not apply it to Task~2, as subjective ratings should be location-independent, while a prior over objective ratings would require training encoders on each ground-truth raster, which we leave to future work.
\section{Experiments}
\label{sec:experiments}

\subsection{Experimental Settings}
\label{sec:setup}

\paragraph{Decision Models.}
We evaluate six decision models in zero-shot settings; four open-weight models: \mpplx{} \citep{perplexity2026decider}, \mglide{} \citep{fastino2026glide}, \mknowline{} \citep{pelaai2026knowline}, and \msieve{} \citep{sthanika2026sieve}, and two closed-source models: \mluna{} (Decisions) \citep{openai2026decisions} and \mjev{} \citep{typesafe2026jev}.
To isolate the effect of adapting base LLMs into decision models, we also evaluate their base models: \mqwenbase{} \citep{qwen38} and \mgemmabase{} \citep{gemmateam2026gemma4}.

\paragraph{Inference.}
We run the open-weight models with their released inference code on all samples.
Since base LLMs have no decision head, we list the candidate labels with letters and take the softmax over the letters' next-token log-probabilities, following standard practice.
Due to API cost, the closed-source models are only evaluated on a fixed 2,000-sample subset of each Task 1 dataset, while all Task 2 samples are included.

\paragraph{TorchSpatial Location Encoders and Baselines.}
For the location prior $g$ in Eq.~\ref{eq:locprior}, we use the pretrained Space2Vec-theory \citep{mai2020space2vec} location encoders released with TorchSpatial \citep{wu2024torchspatial}, which output per-class sigmoid scores at any coordinate.
These encoders were trained on the LocBench training split of each dataset and we use them without further training.
For Task 1, we use the pretrained TorchSpatial image classifiers (InceptionV3 \citep{szegedy2016rethinking} for YFCC and iNat and ResNet-50 \citep{ayush2021geography} for fMoW) as the baseline for the Image setting.
For the Coordinates setting, we use the same pretrained Space2Vec-theory location encoder.
For Image $\times$ Location Prior, we multiply the image classifier's predictions by the corresponding location prior.
For Task 2, we compare our results with the bias scores of generative LLMs reported by \citet{manvi2024large} in the Full setting (Appendix~\ref{app:excess}).

\subsection{Spatial Awareness (RQ1)}
\label{sec:rq1}

Table~\ref{tab:track1_acc} reports top-1 accuracy on Task 1, and Table~\ref{tab:track2_obj} reports Spearman $\rho$ on Task 2 objective topics.

\paragraph{Spatial awareness requires vision.}
Across the Task 1 datasets, the open-weight decision models perform poorly in the Coordinates setting, reaching at most 6.8\% top-1 accuracy on YFCC, 4.4\% on fMoW, 19.0\% on iNat2017, and 18.5\% on iNat2018.
These results are only marginally above random guessing on YFCC and fMoW (1.0\% and 1.6\%), and below the Space2Vec-theory location prior alone on all four datasets.
This suggests that coordinates alone give decision models insufficient spatial context for image classification, even though language models may implicitly encode spatial knowledge \citep{gurnee2024language}.
In contrast, decision models perform significantly better in the Image setting compared to the Coordinates setting.
On YFCC, which consists of everyday photos, the open-weight models reach 62.8--64.4\% and outperform the supervised TorchSpatial image classifier (50.1\%), but this advantage does not extend to the satellite (fMoW) and fine-grained species (iNat) images.
In the Image + Coordinates setting, the open-weight models improve over the Image setting by 5.4--7.0\% on the two iNat datasets, but gain negligibly on YFCC and fMoW.

\paragraph{Human geography requires place names.}
On Task 2, the Coordinates setting allows decision models to rank locations by temperature, precipitation, and infant mortality ($\rho=0.69$--$0.88$), but provides limited spatial context for nighttime lights, population density, and built-up ratio ($\rho\leq 0.40$).
In the Place Name setting, correlations on these human-geography topics increase to $\rho=0.69$--$0.87$.
This gap likely reflects the spatial distribution of the topics: temperature, precipitation, and infant mortality vary smoothly across regions, whereas human geography is spatially heterogeneous at fine scales (e.g., urban vs.\ rural areas) and is tied to named places, which LLMs represent well \citep{gurnee2024language}.
\mluna{} is a notable exception, reaching $\rho\approx 0.7$ on human-geography topics even in the Coordinates setting.

\begin{table*}[!t]
    \centering
    \small
    \setlength{\tabcolsep}{2.6pt}
    \resizebox{\textwidth}{!}{%
        \begin{tabular}{@{}l cccc|cccc|cccc|cccc@{}}
            \toprule
                                      & \multicolumn{4}{c|}{\textbf{YFCC}} & \multicolumn{4}{c|}{\textbf{fMoW}}       & \multicolumn{4}{c|}{\textbf{iNat2017}} & \multicolumn{4}{c@{}}{\textbf{iNat2018}}                                                                                                                                                                                                                                                                                               \\
            \cmidrule(lr){2-5}\cmidrule(lr){6-9}\cmidrule(lr){10-13}\cmidrule(lr){14-17}
                                      & \multicolumn{2}{c}{\textbf{I}}     & \multicolumn{2}{c|}{\textbf{I$\times$P}} & \multicolumn{2}{c}{\textbf{I}}         & \multicolumn{2}{c|}{\textbf{I$\times$P}} & \multicolumn{2}{c}{\textbf{I}} & \multicolumn{2}{c|}{\textbf{I$\times$P}} & \multicolumn{2}{c}{\textbf{I}} & \multicolumn{2}{c@{}}{\textbf{I$\times$P}}                                                                                                                                     \\
            \cmidrule(lr){2-3}\cmidrule(lr){4-5}\cmidrule(lr){6-7}\cmidrule(lr){8-9}\cmidrule(lr){10-11}\cmidrule(lr){12-13}\cmidrule(lr){14-15}\cmidrule(lr){16-17}
            \textbf{Model}            & \ssiex{}                           & \sreex{}                                 & \ssiex{}                               & \sreex{}                                 & \ssiex{}                       & \sreex{}                                 & \ssiex{}                       & \sreex{}                                   & \ssiex{} & \sreex{}         & \ssiex{}         & \sreex{}         & \ssiex{} & \sreex{}      & \ssiex{}        & \sreex{}         \\
            \midrule
            TorchSpatial              & \underline{5.9}                    & \underline{.005}                         & \textbf{5.2}                           & \textbf{.003}                            & \underline{333}                & .109                                     & \textbf{331}                   & .114                                       & 40.1     & .005             & 53.9             & .004             & 23.3     & .007          & 28.4            & .007             \\
            \midrule
            \logo{qwen}\mqwenbase{}   & 15.4                               & .007                                     & 16.4                                   & .009                                     & 336                            & .084                                     & 344                            & \underline{.076}                           & 20.5     & \textbf{.002}    & \underline{12.1} & .005             & 37.2     & .005          & \textbf{4.9}    & \textbf{.003}    \\
            \logo{gemma}\mgemmabase{} & 17.1                               & .009                                     & 16.6                                   & .009                                     & 345                            & \textbf{.074}                            & 343                            & \textbf{.074}                              & 28.5     & \underline{.003} & \textbf{9.1}     & \underline{.003} & 48.9     & .008          & 8.7             & \textbf{.003}    \\
            \midrule
            \logo{qwen}\mpplx{}       & 19.5                               & .010                                     & 22.9                                   & .009                                     & 363                            & .109                                     & 365                            & .090                                       & 16.2     & \textbf{.002}    & 16.3             & .005             & 34.2     & \textbf{.003} & 6.4             & \textbf{.003}    \\
            \logo{qwen}\mglide{}      & 26.6                               & .007                                     & 26.3                                   & .008                                     & 363                            & .129                                     & 365                            & .106                                       & 20.2     & \underline{.003} & 17.2             & .005             & 35.8     & .006          & 7.2             & \underline{.004} \\
            \logo{gemma}\mknowline{}  & 21.9                               & .007                                     & 21.6                                   & .009                                     & 354                            & .089                                     & 354                            & .087                                       & 30.3     & \underline{.003} & 12.5             & .005             & 53.9     & .007          & 7.3             & .005             \\
            \logo{gemma}\msieve{}     & 28.2                               & .008                                     & 26.6                                   & .009                                     & 370                            & .112                                     & 373                            & .102                                       & 23.5     & \underline{.003} & 16.5             & .004             & 47.1     & .006          & \underline{6.3} & \underline{.004} \\
            \bottomrule
        \end{tabular}}
    \caption{Shuffle-corrected geo-bias scores ($\downarrow$) on Task 1 in the Image (I) and Image $\times$ Location Prior (I$\times$P) settings. Bold and underlined values indicate the \textbf{lowest} and \underline{second-lowest} scores per dataset and metric, respectively. Results for all four input settings and raw SSI and SRE values are provided in Appendix~\ref{app:excess}. Closed-source models are omitted, as their 2,000-sample subsets are too sparse for comparable geo-bias scores.}
    \label{tab:track1_geobias}
\end{table*}

\begin{table*}[!t]
\centering
\small
\setlength{\tabcolsep}{2.6pt}
\resizebox{\textwidth}{!}{%
\begin{tabular}{@{}l >{\centering\arraybackslash}p{2.6em}>{\centering\arraybackslash}p{2.6em}>{\centering\arraybackslash}p{2.6em}|>{\centering\arraybackslash}p{2.6em}>{\centering\arraybackslash}p{2.6em}>{\centering\arraybackslash}p{2.6em}|>{\centering\arraybackslash}p{2.6em}>{\centering\arraybackslash}p{2.6em}>{\centering\arraybackslash}p{2.6em}|>{\centering\arraybackslash}p{2.6em}>{\centering\arraybackslash}p{2.6em}>{\centering\arraybackslash}p{2.6em}|>{\centering\arraybackslash}p{2.6em}>{\centering\arraybackslash}p{2.6em}>{\centering\arraybackslash}p{2.6em}@{}}
\toprule
 & \multicolumn{3}{c|}{\textbf{Intelligence}} & \multicolumn{3}{c|}{\textbf{Work Ethic}} & \multicolumn{3}{c|}{\textbf{Morality}} & \multicolumn{3}{c|}{\textbf{Likability}} & \multicolumn{3}{c@{}}{\textbf{Attractiveness}} \\
\cmidrule(lr){2-4}\cmidrule(lr){5-7}\cmidrule(lr){8-10}\cmidrule(lr){11-13}\cmidrule(lr){14-16}
\textbf{Model} & \textbf{C} & \textbf{P} & \textbf{F} & \textbf{C} & \textbf{P} & \textbf{F} & \textbf{C} & \textbf{P} & \textbf{F} & \textbf{C} & \textbf{P} & \textbf{F} & \textbf{C} & \textbf{P} & \textbf{F} \\
\midrule
\logo{qwen}\mqwenbase{} & $-$0.49 & $-$0.61 & $-$0.50 & $-$0.37 & $-$0.59 & $-$0.47 & $-$0.17 & $-$0.25 & $-$0.26 & $-$0.15 & $-$0.20 & $-$0.17 & $-$0.07 & $-$0.09 & $-$0.19 \\
\logo{gemma}\mgemmabase{} & $-$0.65 & $-$0.80 & $-$0.73 & $-$0.38 & $-$0.48 & $-$0.48 & \textbf{0.01} & $-$0.04 & $-$0.09 & \textbf{0.00} & $-$0.07 & $-$0.08 & \textbf{0.00} & 0.03 & \underline{0.01} \\
\midrule
\logo{qwen}\mpplx{} & $-$0.44 & $-$0.57 & $-$0.51 & $-$0.34 & $-$0.47 & $-$0.37 & $-$0.32 & $-$0.38 & $-$0.32 & $-$0.15 & $-$0.19 & $-$0.17 & $-$0.18 & $-$0.18 & $-$0.11 \\
\logo{qwen}\mglide{} & $-$0.41 & $-$0.50 & $-$0.45 & $-$0.32 & $-$0.42 & $-$0.33 & $-$0.22 & $-$0.32 & $-$0.28 & $-$0.21 & $-$0.21 & $-$0.19 & $-$0.22 & $-$0.17 & $-$0.13 \\
\logo{gemma}\mknowline{} & $-$0.75 & $-$0.95 & $-$0.84 & $-$0.61 & $-$0.81 & $-$0.79 & $-$0.43 & $-$0.45 & $-$0.41 & $-$0.12 & $-$0.20 & $-$0.23 & $-$0.21 & $-$0.34 & $-$0.33 \\
\logo{gemma}\msieve{} & $-$0.55 & $-$0.72 & $-$0.68 & $-$0.33 & $-$0.46 & $-$0.47 & $-$0.21 & $-$0.25 & $-$0.26 & $-$0.13 & $-$0.22 & $-$0.30 & $-$0.19 & $-$0.39 & $-$0.33 \\
\midrule
\logo{openai}\mluna{} & \textbf{$-$0.04} & \underline{$-$0.05} & \underline{$-$0.05} & $-$0.25 & $-$0.24 & \textbf{$-$0.17} & \underline{$-$0.02} & \textbf{$-$0.01} & \textbf{$-$0.01} & $-$0.09 & \underline{$-$0.05} & $-$0.06 & \underline{$-$0.01} & \underline{$-$0.01} & \underline{$-$0.01} \\
\logo{typesafe}\mjev{} & $-$0.43 & $-$0.63 & $-$0.64 & \underline{$-$0.23} & $-$0.26 & $-$0.28 & $-$0.16 & $-$0.19 & $-$0.18 & $-$0.13 & $-$0.14 & $-$0.13 & $-$0.15 & $-$0.19 & $-$0.19 \\
\bottomrule
\end{tabular}}
\caption{Bias score $B$ ($\rightarrow 0$) on the subjective topics of Task 2 (intelligence, work ethic, morality, likability, and attractiveness) in the Coordinates (C), Place Name (P), and Full (F) settings; $B<0$ means lower ratings in regions with lower socioeconomic conditions. Bold and underlined values indicate the \textbf{least} and \underline{second-least} biased results per topic, respectively.}
\label{tab:track2_subj}
\end{table*}

\paragraph{Decision fine-tuning brings modest gains.}
Comparing open-weight decision models with their base LLMs reveals that adapting LLMs into decision models yields modest gains in spatial awareness.
On Task~1, decision fine-tuning brings marginal improvements in the Image setting, while in the Coordinates setting, only the Qwen-based models consistently outperform their base LLM.
On Task~2, it improves rank correlations for temperature, precipitation, and infant mortality in the Coordinates setting, but has minimal impact in the Place Name and Full settings.
Overall, decision fine-tuning mainly improves how models leverage coordinates as textual inputs, while most of their spatial knowledge comes from pretraining.

\subsection{Geographic Bias (RQ2)}
\label{sec:rq2}

Table~\ref{tab:track1_geobias} reports the shuffle-corrected geo-bias scores on Task 1, and Table~\ref{tab:track2_subj} reports the bias scores on the subjective topics of Task 2.

\paragraph{Decision models exhibit spatially clustered errors.}
On YFCC, decision models outperform TorchSpatial image classifiers in terms of accuracy, but their errors are 3--5$\times$ more spatially clustered in the Image setting (\ssiex{} 19.5--28.2 vs.\ 5.9), potentially reflecting regional imbalances in web pretraining data \citep{devries2019does}.
On iNat2017, adding Coordinates to the Image lowers \ssiex{} for all open-weight decision models (e.g., \mglide{}: 20.2$\rightarrow$13.4), unlike TorchSpatial models whose errors become more spatially clustered when a location prior is added \citep{wu2024torchspatial}.
Geo-bias scores must be read together with accuracy; in the Coordinates setting, the decision models are only marginally above random guessing on YFCC and fMoW (Section~\ref{sec:rq1}), so most predictions are errors and their geo-bias scores are uninformative.
On iNat, where accuracy in the Coordinates setting is substantially higher, Qwen-based decision models exhibit strongly clustered errors (\ssiex{} 69.3--93.2), indicating a strong geographic \textit{performance} bias.

\paragraph{Decision models exhibit social bias.}
On all five subjective topics and in every input setting, all decision models give lower ratings in regions with lower socioeconomic conditions ($B<0$).
In the Full setting, intelligence is the most biased topic for every model except \mluna{}.
On intelligence, every model (except \mluna{}) is more biased than \mgptthreefive{}.
\mknowline{}, \msieve{}, and \mjev{} are also more biased than \mgeminipro{}, the most biased generative LLM reported by \citet{manvi2024large} (Appendix~\ref{app:excess}).
This bias is largely inherited from the base LLMs, although decision fine-tuning may add bias on some topics. For instance, the base \mgemmabase{} is nearly unbiased on attractiveness and morality, whereas both Gemma-based decision models exhibit a stronger bias on these topics.
\mluna{} is an exception: its $B$ is near zero on intelligence, attractiveness, and morality because it rates almost every location about 5.0 on these topics, yet its intelligence ratings remain correlated with infant mortality, and it refuses to answer 8.8\% of intelligence questions, whereas \mjev{} refuses none.
A near-zero $B$ can thus hide a biased ranking and should be read together with its correlation with socioeconomic conditions.

\subsection{Spatial Awareness with a Location Prior (RQ3)}
\label{sec:rq3}

Tables~\ref{tab:track1_acc} and~\ref{tab:track1_geobias} report the Image $\times$ Location Prior setting, which combines decision models with the Space2Vec-theory location prior (Section~\ref{sec:method}).

\paragraph{Location priors enable spatial awareness.}
Our training-free method improves the accuracy of decision models by combining their predictions with a pretrained location prior.
On fMoW and iNat, every decision model with the location prior outperforms both the Image setting and the TorchSpatial location encoder, with accuracy gains of +3.8--6.0\% on fMoW, +12.9--14.0\% on iNat2017, and +25.2--26.7\% on iNat2018 for the open-weight models.
The method also generalizes to closed-source models, with gains of up to +25.5\% for \mjev{} on iNat2018.
With the same prior, every open-weight decision model outperforms its base LLM on both iNat datasets.
Consistent with RQ1, decision models cannot infer spatial context from textual coordinates alone, but can leverage the spatial information provided by a location prior.
On YFCC, where geographic location is less informative for everyday photos, neither coordinates nor the location prior improves accuracy, although decision models still outperform the supervised TorchSpatial baseline.

\paragraph{Location priors do not substantially increase geo-bias.}
Unlike the supervised TorchSpatial image classifier, whose errors on iNat become more spatially clustered with a location prior \citep{wu2024torchspatial}, adding a location prior does not substantially increase the spatial clustering of errors in the open-weight decision models.
On iNat2018, the location prior substantially reduces \ssiex{} for all four open-weight decision models (34.2--53.9 $\rightarrow$ 6.3--7.3) and both base LLMs (37.2--48.9 $\rightarrow$ 4.9--8.7), well below the supervised baseline with the same prior.
On iNat2017, \ssiex{} changes by at most $+$0.1 over the Image setting and stays below 18, whereas the supervised baseline rises from 40.1 to 53.9. \sreex{} rises only slightly from .002--.003 to .004--.005.
Similarly, on fMoW, the location prior lowers \sreex{} for all four open-weight decision models, in contrast to the increase observed for the supervised baseline.
Adding a location prior therefore improves the spatial awareness of decision models without the substantial increase in performance bias observed in supervised models.
\section{Conclusion}
\label{sec:conclusion}

We introduced \dataset{}, a multimodal benchmark for evaluating the spatial awareness and geographic bias of decision models.
Across six decision models, we found that spatial awareness requires visual context, and that textual coordinates alone are insufficient for classifying everyday photos and satellite images.
For location rating, coordinates allow decision models to rank locations by temperature, precipitation, and infant mortality, but place names are needed to capture human geography.
We also found that decision models exhibit substantial geographic bias: the errors of open-weight decision models on everyday photos are more spatially clustered than those of supervised models, despite higher accuracy, and all six models give lower ratings to locations in regions with lower socioeconomic conditions.
This bias is largely inherited from the base LLMs, but decision fine-tuning can further increase bias on some topics.
Finally, our training-free method, which combines decision models with a pretrained location prior, improves spatial awareness without substantially increasing the performance bias of decision models.
We release \dataset{} and our code to support future research on spatially aware and geographically fair decision models.
\section*{Limitations}

\paragraph{Evaluation Setup.}
Following \citet{manvi2024large}, we use infant mortality as the only proxy for socioeconomic conditions. Other proxies, such as population density, may capture different forms of social bias.
Furthermore, our location prior uses TorchSpatial location encoders trained on each LocBench training split, making the prior dataset-specific. We therefore apply it only to Task~1 and leave its extension to Task~2 for future work.

\paragraph{Comparability across Models.}
Image preprocessing varies across models: each open-weight model uses its own image processor, while \mknowline{} receives images resized to fit \mjev{}'s input size limits. Consequently, approximately 50\% of fMoW images and 4\% of YFCC images are upsampled for \mknowline{}.
We also evaluate \mknowline{} and \msieve{} using their released inference configurations, which differ from those of other models by design. These include repeated prompting and task-specific temperatures for \mknowline{}, and two label orders and category-specific temperatures for \msieve{}.
Finally, \mjev{} and \mluna{} predict ratings on the 100-level scale through classification over probabilities rounded to two decimal places, resulting in coarser predicted ratings than those of open-weight models.

\section*{Ethical Considerations}

Task~2 asks models to rate the intelligence, morality, attractiveness, likability, and work ethic of residents of real places. We include these questions solely to measure social bias, following \citet{manvi2024large}, and use infant mortality only as a proxy for socioeconomic conditions. We do not endorse any of the resulting ratings.
All datasets are public research benchmarks used under their original licenses. We release our annotations, model predictions, and the source data whose licenses permit redistribution for non-commercial research; iNaturalist images are not redistributed and must be obtained from their original source.

\bibliography{custom}

\begin{thebibliography}{56}
\providecommand{\natexlab}[1]{#1}

\bibitem[{Abouelhadid et~al.(2026)Abouelhadid, Abuhani, Munir, Khan, and
  Khan}]{abouelhadid2026eojev}
Mohamed~Atef Abouelhadid, Diaa~Addeen Abuhani, Muhammad~Akhtar Munir,
  Fahad~Shahbaz Khan, and Salman Khan. 2026.
\newblock \href {https://arxiv.org/abs/2611.03371} {{EO-Jev}: Decision model
  for earth observation}.
\newblock \emph{Preprint}, arXiv:2611.03371.

\bibitem[{Allahyar et~al.(2026)Allahyar, White, Atreja, Hurn-Maloney, and
  Zaratiana}]{fastino2026glide}
Sahibzada Allahyar, Julia White, Dhruv Atreja, George Hurn-Maloney, and Urchade
  Zaratiana. 2026.
\newblock \href {https://doi.org/10.48550/arXiv.2611.26509} {Glide: A fast,
  calibrated decision model distilled from reasoning teachers}.
\newblock arXiv preprint arXiv:2611.26509.
\newblock \emph{arXiv preprint}.

\bibitem[{Ayush et~al.(2021)Ayush, Uzkent, Meng, Tanmay, Burke, Lobell, and
  Ermon}]{ayush2021geography}
Kumar Ayush, Burak Uzkent, Chenlin Meng, Kumar Tanmay, Marshall Burke, David
  Lobell, and Stefano Ermon. 2021.
\newblock \href {https://doi.org/10.1109/iccv48922.2021.01002} {Geography-aware
  self-supervised learning}.
\newblock In \emph{2021 IEEE/CVF International Conference on Computer Vision
  (ICCV)}, pages 10161--10170. IEEE.

\bibitem[{Balloccu et~al.(2024)Balloccu, Schmidtov{\'a}, Lango, and
  Dusek}]{balloccu2024leak}
Simone Balloccu, Patr{\'i}cia Schmidtov{\'a}, Mateusz Lango, and Ondrej Dusek.
  2024.
\newblock \href {https://doi.org/10.18653/v1/2024.eacl-long.5} {Leak, cheat,
  repeat: Data contamination and evaluation malpractices in closed-source
  {LLM}s}.
\newblock In \emph{Proceedings of the 18th Conference of the {E}uropean Chapter
  of the {A}ssociation for {C}omputational {L}inguistics (Volume 1: Long
  Papers)}, pages 67--93, St. Julian{'}s, Malta. Association for Computational
  Linguistics.

\bibitem[{Christie et~al.(2018)Christie, Fendley, Wilson, and
  Mukherjee}]{christie2018fmow}
Gordon Christie, Neil Fendley, James Wilson, and Ryan Mukherjee. 2018.
\newblock \href {https://doi.org/10.1109/cvpr.2018.00646} {Functional map of
  the world}.
\newblock In \emph{2018 IEEE/CVF Conference on Computer Vision and Pattern
  Recognition}, pages 6172--6180. IEEE.

\bibitem[{{CIESIN}(2021)}]{ciesin2021imr}
{CIESIN}. 2021.
\newblock \href {https://doi.org/10.7927/0gdn-6y33} {Global subnational infant
  mortality rates, version 2.01}.
\newblock NASA Socioeconomic Data and Applications Center (SEDAC), Palisades,
  NY.

\bibitem[{Corbane et~al.(2019)Corbane, Pesaresi, Kemper, Politis, Florczyk,
  Syrris, Melchiorri, Sabo, and Soille}]{corbane2019ghsl}
Christina Corbane, Martino Pesaresi, Thomas Kemper, Panagiotis Politis,
  Aneta~J. Florczyk, Vasileios Syrris, Michele Melchiorri, Filip Sabo, and
  Pierre Soille. 2019.
\newblock \href {https://doi.org/10.1080/20964471.2019.1625528} {Automated
  global delineation of human settlements from 40 years of {Landsat} satellite
  data archives}.
\newblock \emph{Big Earth Data}, 3(2):140--169.

\bibitem[{de~Vries et~al.(2019)de~Vries, Misra, Wang, and van~der
  Maaten}]{devries2019does}
Terrance de~Vries, Ishan Misra, Changhan Wang, and Laurens van~der Maaten.
  2019.
\newblock Does object recognition work for everyone?
\newblock In \emph{Proceedings of the IEEE/CVF Conference on Computer Vision
  and Pattern Recognition (CVPR) Workshops}.

\bibitem[{DeepSeek-AI et~al.(2024)DeepSeek-AI, Liu, Feng, Xue, Wang, Wu, Lu,
  Zhao, Deng, Zhang, Ruan, Dai, Guo, Yang, Chen, Ji, Li, Lin, Dai, Luo, Hao,
  Chen, Li, Zhang, Bao, Xu, Wang, Zhang, Ding, Xin, Gao, Li, Qu, Cai, Liang,
  Guo, Ni, Li, Wang, Chen, Chen, Yuan, Qiu, Li, Song, Dong, Hu, Gao, Guan,
  Huang, Yu, Wang, Zhang, Xu, Xia, Zhao, Wang, Zhang, Li, Wang, Zhang, Zhang,
  Tang, Li, Tian, Huang, Wang, Zhang, Wang, Zhu, Chen, Du, Chen, Jin, Ge,
  Zhang, Pan, Wang, Xu, Zhang, Chen, Li, Lu, Zhou, Chen, Wu, Ye, Ye, Ma, Wang,
  Zhou, Yu, Zhou, Pan, Wang, Yun, Pei, Sun, Xiao, Zeng, Zhao, An, Liu, Liang,
  Gao, Yu, Zhang, Li, Jin, Wang, Bi, Liu, Wang, Shen, Chen, Zhang, Chen, Nie,
  Sun, Wang, Cheng, Liu, Xie, Liu, Yu, Song, Shan, Zhou, Yang, Li, Su, Lin, Li,
  Wang, Wei, Zhu, Zhang, Xu, Xu, Huang, Li, Zhao, Sun, Li, Wang, Yu, Zheng,
  Zhang, Shi, Xiong, He, Tang, Piao, Wang, Tan, Ma, Liu, Guo, Wu, Ou, Zhu,
  Wang, Gong, Zou, He, Zha, Xiong, Ma, Yan, Luo, You, Liu, Zhou, Wu, Ren, Ren,
  Sha, Fu, Xu, Huang, Zhang, Xie, Zhang, Hao, Gou, Ma, Yan, Shao, Xu, Wu,
  Zhang, Li, Gu, Zhu, Liu, Li, Xie, Song, Gao, and
  Pan}]{deepseekai2024deepseek}
DeepSeek-AI, Aixin Liu, Bei Feng, Bing Xue, Bingxuan Wang, Bochao Wu, Chengda
  Lu, Chenggang Zhao, Chengqi Deng, Chenyu Zhang, Chong Ruan, Damai Dai, Daya
  Guo, Dejian Yang, Deli Chen, Dongjie Ji, Erhang Li, Fangyun Lin, Fucong Dai,
  and 181 others. 2024.
\newblock \href {https://doi.org/10.48550/arXiv.2412.19437} {{DeepSeek-V3}
  technical report}.
\newblock \emph{arXiv preprint arXiv:2412.19437}.

\bibitem[{Dhamala et~al.(2021)Dhamala, Sun, Kumar, Krishna, Pruksachatkun,
  Chang, and Gupta}]{dhamala2021bold}
Jwala Dhamala, Tony Sun, Varun Kumar, Satyapriya Krishna, Yada Pruksachatkun,
  Kai-Wei Chang, and Rahul Gupta. 2021.
\newblock \href {https://doi.org/10.1145/3442188.3445924} {{BOLD}: Dataset and
  metrics for measuring biases in open-ended language generation}.
\newblock In \emph{Proceedings of the 2021 ACM Conference on Fairness,
  Accountability, and Transparency}, pages 862--872. ACM.

\bibitem[{Elvidge et~al.(2021)Elvidge, Zhizhin, Ghosh, Hsu, and
  Taneja}]{elvidge2021viirs}
Christopher~D. Elvidge, Mikhail Zhizhin, Tilottama Ghosh, Feng-Chi Hsu, and Jay
  Taneja. 2021.
\newblock \href {https://doi.org/10.3390/rs13050922} {Annual time series of
  global {VIIRS} nighttime lights derived from monthly averages: 2012 to 2019}.
\newblock \emph{Remote Sensing}, 13(5):922.

\bibitem[{Faisal and Anastasopoulos(2023)}]{faisal2023geographic}
Fahim Faisal and Antonios Anastasopoulos. 2023.
\newblock \href {https://doi.org/10.18653/v1/2023.mrl-1.12} {Geographic and
  geopolitical biases of language models}.
\newblock In \emph{Proceedings of the 3rd Workshop on Multi-lingual
  Representation Learning ({MRL})}, pages 139--163, Singapore. Association for
  Computational Linguistics.

\bibitem[{Gao et~al.(2026)Gao, Li, Li, Chang, and Wu}]{gao2026ordinal}
Tianxiang Gao, Jinzhe Li, Zhiyuan Li, Yi~Chang, and Yuan Wu. 2026.
\newblock More choices, fewer decisions: Ordinal-scale bias in {JEV-like}
  direct-decision models.
\newblock \emph{arXiv preprint arXiv:2609.38827}.

\bibitem[{{Gemma Team}(2026)}]{gemmateam2026gemma4}
{Gemma Team}. 2026.
\newblock \href {https://arxiv.org/abs/2607.02770} {Gemma 4 technical report}.
\newblock \emph{Preprint}, arXiv:2607.02770.

\bibitem[{Golchin and Surdeanu(2025)}]{golchin2023quiz}
Shahriar Golchin and Mihai Surdeanu. 2025.
\newblock Data contamination quiz: A tool to detect and estimate contamination
  in large language models.
\newblock \emph{Transactions of the Association for Computational Linguistics},
  13:809--830.

\bibitem[{Gurnee and Tegmark(2024)}]{gurnee2024language}
Wes Gurnee and Max Tegmark. 2024.
\newblock \href
  {https://proceedings.iclr.cc/paper_files/paper/2024/file/0a6059857ae5c82ea9726ee9282a7145-Paper-Conference.pdf}
  {Language models represent space and time}.
\newblock In \emph{International Conference on Learning Representations},
  volume 2024, pages 2483--2503.

\bibitem[{Holtzman et~al.(2021)Holtzman, West, Shwartz, Choi, and
  Zettlemoyer}]{holtzman2021surface}
Ari Holtzman, Peter West, Vered Shwartz, Yejin Choi, and Luke Zettlemoyer.
  2021.
\newblock \href {https://doi.org/10.18653/v1/2021.emnlp-main.564} {Surface form
  competition: Why the highest probability answer isn{'}t always right}.
\newblock In \emph{Proceedings of the 2021 Conference on Empirical Methods in
  Natural Language Processing}, pages 7038--7051, Online and Punta Cana,
  Dominican Republic. Association for Computational Linguistics.

\bibitem[{Karger et~al.(2017)Karger, Conrad, B{\"o}hner, Kawohl, Kreft,
  Soria-Auza, Zimmermann, Linder, and Kessler}]{karger2017chelsa}
Dirk~Nikolaus Karger, Olaf Conrad, J{\"u}rgen B{\"o}hner, Tobias Kawohl, Holger
  Kreft, Rodrigo~Wilber Soria-Auza, Niklaus~E. Zimmermann, H.~Peter Linder, and
  Michael Kessler. 2017.
\newblock \href {https://doi.org/10.1038/sdata.2017.122} {Climatologies at high
  resolution for the {Earth}'s land surface areas}.
\newblock \emph{Scientific Data}, 4:170122.

\bibitem[{Li et~al.(2026)Li, Miao, Krishnan, and Padman}]{li2026jevjudge}
Yubo Li, Yidi Miao, Ramayya Krishnan, and Rema Padman. 2026.
\newblock {JEV-as-a-Judge}: Accept when confident, escalate when unsure.
\newblock \emph{arXiv preprint arXiv:2609.26550}.

\bibitem[{Mac~Aodha et~al.(2019)Mac~Aodha, Cole, and
  Perona}]{macaodha2019presence}
Oisin Mac~Aodha, Elijah Cole, and Pietro Perona. 2019.
\newblock \href {https://doi.org/10.1109/iccv.2019.00969} {Presence-only
  geographical priors for fine-grained image classification}.
\newblock In \emph{2019 IEEE/CVF International Conference on Computer Vision
  (ICCV)}, pages 9595--9605. IEEE.

\bibitem[{Mai et~al.(2022)Mai, Janowicz, Hu, Gao, Yan, Zhu, Cai, and
  Lao}]{mai2022review}
Gengchen Mai, Krzysztof Janowicz, Yingjie Hu, Song Gao, Bo~Yan, Rui Zhu, Ling
  Cai, and Ni~Lao. 2022.
\newblock \href {https://doi.org/10.1080/13658816.2021.2004602} {A review of
  location encoding for {GeoAI}: methods and applications}.
\newblock \emph{International Journal of Geographical Information Science},
  36(4):639--673.

\bibitem[{Mai et~al.(2020)Mai, Janowicz, Yan, Zhu, Cai, and
  Lao}]{mai2020space2vec}
Gengchen Mai, Krzysztof Janowicz, Bo~Yan, Rui Zhu, Ling Cai, and Ni~Lao. 2020.
\newblock Multi-scale representation learning for spatial feature distributions
  using grid cells.
\newblock In \emph{International Conference on Learning Representations}.
  openreview.

\bibitem[{Mai et~al.(2023)Mai, Xuan, Zuo, He, Song, Ermon, Janowicz, and
  Lao}]{mai2023sphere2vec}
Gengchen Mai, Yao Xuan, Wenyun Zuo, Yutong He, Jiaming Song, Stefano Ermon,
  Krzysztof Janowicz, and Ni~Lao. 2023.
\newblock \href {https://doi.org/10.1016/j.isprsjprs.2023.06.016}
  {{Sphere2Vec}: A general-purpose location representation learning over a
  spherical surface for large-scale geospatial predictions}.
\newblock \emph{ISPRS Journal of Photogrammetry and Remote Sensing},
  202:439--462.

\bibitem[{Manvi et~al.(2024{\natexlab{a}})Manvi, Khanna, Burke, Lobell, and
  Ermon}]{manvi2024large}
Rohin Manvi, Samar Khanna, Marshall Burke, David Lobell, and Stefano Ermon.
  2024{\natexlab{a}}.
\newblock Large language models are geographically biased.
\newblock In \emph{Proceedings of the 41st International Conference on Machine
  Learning}, ICML'24. JMLR.org.

\bibitem[{Manvi et~al.(2024{\natexlab{b}})Manvi, Khanna, Mai, Burke, Lobell,
  and Ermon}]{manvi2024geollm}
Rohin Manvi, Samar Khanna, Gengchen Mai, Marshall Burke, David~B. Lobell, and
  Stefano Ermon. 2024{\natexlab{b}}.
\newblock \href {https://openreview.net/forum?id=TqL2xBwXP3} {Geo{LLM}:
  Extracting geospatial knowledge from large language models}.
\newblock In \emph{The Twelfth International Conference on Learning
  Representations}.

\bibitem[{Moayeri et~al.(2024)Moayeri, Tabassi, and
  Feizi}]{moayeri2024worldbench}
Mazda Moayeri, Elham Tabassi, and Soheil Feizi. 2024.
\newblock \href {https://doi.org/10.1145/3630106.3658967} {{WorldBench}:
  Quantifying geographic disparities in {LLM} factual recall}.
\newblock In \emph{The 2024 ACM Conference on Fairness, Accountability, and
  Transparency}, pages 1211--1228. ACM.

\bibitem[{Niu et~al.(2026)Niu, Chen, Zhang, Hu, Liu, Liu, and
  Liu}]{niu2026geojev}
Xinyi Niu, Lingfeng Chen, Jiehua Zhang, Panhe Hu, Li~Liu, Zhen Liu, and
  Yongxiang Liu. 2026.
\newblock \href {https://arxiv.org/abs/2611.08879} {{GeoJeV}: Unified spatial
  probability for training-free efficient evidence acquisition in
  ultra-high-resolution remote sensing}.
\newblock \emph{Preprint}, arXiv:2611.08879.

\bibitem[{Nwatu et~al.(2023)Nwatu, Ignat, and Mihalcea}]{nwatu2023bridging}
Joan Nwatu, Oana Ignat, and Rada Mihalcea. 2023.
\newblock \href {https://doi.org/10.18653/v1/2023.emnlp-main.660} {Bridging the
  digital divide: Performance variation across socio-economic factors in
  vision-language models}.
\newblock In \emph{Proceedings of the 2023 Conference on Empirical Methods in
  Natural Language Processing}, pages 10686--10702, Singapore. Association for
  Computational Linguistics.

\bibitem[{{OpenAI}(2026)}]{openai2026decisions}
{OpenAI}. 2026.
\newblock \href {https://developers.openai.com/api/docs/guides/decisions}
  {Decisions {API} guide}.
\newblock Model gpt-6-luna. Accessed 2026-11-04.

\bibitem[{Palmer(2026)}]{palmer2026kev}
Jared Palmer. 2026.
\newblock {Kev-27B}: a calibrated decision model on {Qwen3.8-27B}.
\newblock \url{https://huggingface.co/jaredpalmer/kev-27b}.
\newblock Version 2, released 2026-09-30.

\bibitem[{Parrish et~al.(2022)Parrish, Chen, Nangia, Padmakumar, Phang,
  Thompson, Htut, and Bowman}]{parrish2022bbq}
Alicia Parrish, Angelica Chen, Nikita Nangia, Vishakh Padmakumar, Jason Phang,
  Jana Thompson, Phu~Mon Htut, and Samuel~R. Bowman. 2022.
\newblock \href {https://doi.org/10.18653/v1/2022.findings-acl.165} {{BBQ}: A
  hand-built bias benchmark for question answering}.
\newblock In \emph{Findings of the {A}ssociation for {C}omputational
  {L}inguistics: {ACL} 2022}, pages 2086--2105, Dublin, Ireland. Association
  for Computational Linguistics.

\bibitem[{PelaAI(2026)}]{pelaai2026knowline}
PelaAI. 2026.
\newblock \href {https://huggingface.co/PelaAI/KnowLine-31B-Gen1}
  {{KnowLine-31B-Gen1}: a {31B} decision model}.

\bibitem[{{Perplexity AI}(2026)}]{perplexity2026decider}
{Perplexity AI}. 2026.
\newblock \href {https://huggingface.co/perplexity-ai/pplx-decider-v1.2}
  {pplx-decider-v1.2}.

\bibitem[{Pezeshkpour and Hruschka(2024)}]{pezeshkpour2024large}
Pouya Pezeshkpour and Estevam Hruschka. 2024.
\newblock \href {https://doi.org/10.18653/v1/2024.findings-naacl.130} {Large
  language models sensitivity to the order of options in multiple-choice
  questions}.
\newblock In \emph{Findings of the {A}ssociation for {C}omputational
  {L}inguistics: {NAACL} 2024}, pages 2006--2017, Mexico City, Mexico.
  Association for Computational Linguistics.

\bibitem[{{Qwen Team}(2026)}]{qwen38}
{Qwen Team}. 2026.
\newblock \href {https://qwen.ai/blog?id=qwen3.8} {{Qwen3.8-Max}: A new bar for
  coding and cowork}.

\bibitem[{Rao and Callison-Burch(2026)}]{rao2026jev}
Delip Rao and Chris Callison-Burch. 2026.
\newblock Decision models vs. {LLMs} as rubric judges: Cheaper, faster, and
  wrong in the same places.
\newblock \emph{arXiv preprint arXiv:2609.29769}.

\bibitem[{Robinson and Wingate(2023)}]{robinson2023leveraging}
Joshua Robinson and David Wingate. 2023.
\newblock \href {https://openreview.net/forum?id=yKbprarjc5B} {Leveraging large
  language models for multiple choice question answering}.
\newblock In \emph{The Eleventh International Conference on Learning
  Representations}.

\bibitem[{Sainz et~al.(2023)Sainz, Campos, Garc{\'i}a-Ferrero, Etxaniz,
  de~Lacalle, and Agirre}]{sainz2023nlp}
Oscar Sainz, Jon Campos, Iker Garc{\'i}a-Ferrero, Julen Etxaniz, Oier~Lopez
  de~Lacalle, and Eneko Agirre. 2023.
\newblock \href {https://doi.org/10.18653/v1/2023.findings-emnlp.722} {{NLP}
  evaluation in trouble: On the need to measure {LLM} data contamination for
  each benchmark}.
\newblock In \emph{Findings of the {A}ssociation for {C}omputational
  {L}inguistics: {EMNLP} 2023}, pages 10776--10787, Singapore. Association for
  Computational Linguistics.

\bibitem[{Seshadri et~al.(2022)Seshadri, Pezeshkpour, and
  Singh}]{seshadri2022quantifying}
Preethi Seshadri, Pouya Pezeshkpour, and Sameer Singh. 2022.
\newblock Quantifying social biases using templates is unreliable.
\newblock \emph{arXiv preprint arXiv:2210.04337}.

\bibitem[{Shankar et~al.(2017)Shankar, Halpern, Breck, Atwood, Wilson, and
  Sculley}]{shankar2017no}
Shreya Shankar, Yoni Halpern, Eric Breck, James Atwood, Jimbo Wilson, and
  D.~Sculley. 2017.
\newblock No classification without representation: Assessing geodiversity
  issues in open data sets for the developing world.
\newblock In \emph{NIPS 2017 workshop: Machine Learning for the Developing
  World}.

\bibitem[{Song et~al.(2025)Song, Lai, Wang, Chen, Sun, and
  Wang}]{song2024mmdetect}
Dingjie Song, Sicheng Lai, Mingxuan Wang, Shunian Chen, Lichao Sun, and Benyou
  Wang. 2025.
\newblock \href {https://doi.org/10.18653/v1/2025.findings-emnlp.556} {Both
  text and images leaked! a systematic analysis of data contamination in
  multimodal {LLM}}.
\newblock In \emph{Findings of the {A}ssociation for {C}omputational
  {L}inguistics: {EMNLP} 2025}, pages 10527--10542, Suzhou, China. Association
  for Computational Linguistics.

\bibitem[{{Sthanika AI}(2026)}]{sthanika2026sieve}
{Sthanika AI}. 2026.
\newblock \href {https://github.com/sthanika-ai/Sieve} {Sieve: Decision
  models}.

\bibitem[{Stureborg et~al.(2024)Stureborg, Alikaniotis, and
  Suhara}]{stureborg2024large}
Rickard Stureborg, Dimitris Alikaniotis, and Yoshi Suhara. 2024.
\newblock Large language models are inconsistent and biased evaluators.
\newblock \emph{arXiv preprint arXiv:2405.01724}.

\bibitem[{Szegedy et~al.(2016)Szegedy, Vanhoucke, Ioffe, Shlens, and
  Wojna}]{szegedy2016rethinking}
Christian Szegedy, Vincent Vanhoucke, Sergey Ioffe, Jon Shlens, and Zbigniew
  Wojna. 2016.
\newblock \href {https://doi.org/10.1109/cvpr.2016.308} {Rethinking the
  {Inception} architecture for computer vision}.
\newblock In \emph{2016 IEEE Conference on Computer Vision and Pattern
  Recognition (CVPR)}, pages 2818--2826. IEEE.

\bibitem[{Tang et~al.(2015)Tang, Paluri, Fei-Fei, Fergus, and
  Bourdev}]{tang2015improving}
Kevin Tang, Manohar Paluri, Li~Fei-Fei, Rob Fergus, and Lubomir Bourdev. 2015.
\newblock \href {https://doi.org/10.1109/iccv.2015.121} {Improving image
  classification with location context}.
\newblock In \emph{2015 IEEE International Conference on Computer Vision
  (ICCV)}, pages 1008--1016. IEEE.

\bibitem[{{TypeSafe}(2026)}]{typesafe2026jev}
{TypeSafe}. 2026.
\newblock \href {https://docs.typesafe.ai/api} {{System One} {API} reference}.
\newblock Accessed 2026-11-03.

\bibitem[{Vafa et~al.(2024)Vafa, Chen, Rambachan, Kleinberg, and
  Mullainathan}]{vafa2024evaluating}
Keyon Vafa, Justin Chen, Ashesh Rambachan, Jon Kleinberg, and Sendhil
  Mullainathan. 2024.
\newblock \href {https://doi.org/10.52202/079017-0846} {Evaluating the world
  model implicit in a generative model}.
\newblock In \emph{Advances in Neural Information Processing Systems 37}, pages
  26941--26975.

\bibitem[{Van~Horn et~al.(2018)Van~Horn, Mac~Aodha, Song, Cui, Sun, Shepard,
  Adam, Perona, and Belongie}]{vanhorn2018inaturalist}
Grant Van~Horn, Oisin Mac~Aodha, Yang Song, Yin Cui, Chen Sun, Alex Shepard,
  Hartwig Adam, Pietro Perona, and Serge Belongie. 2018.
\newblock \href {https://doi.org/10.1109/cvpr.2018.00914} {The {iNaturalist}
  species classification and detection dataset}.
\newblock In \emph{2018 IEEE/CVF Conference on Computer Vision and Pattern
  Recognition}, pages 8769--8778. IEEE.

\bibitem[{Wang et~al.(2025{\natexlab{a}})Wang, Zou, Jiang, Wen, Wei, Wen, and
  Liang}]{wang2025urbanr1}
Qiongyan Wang, Xingchen Zou, Yutian Jiang, Haomin Wen, Jiaheng Wei, Qingsong
  Wen, and Yuxuan Liang. 2025{\natexlab{a}}.
\newblock {Urban-R1}: Reinforced {MLLMs} mitigate geospatial biases for urban
  general intelligence.
\newblock \emph{arXiv preprint arXiv:2510.16555}.

\bibitem[{Wang et~al.(2024)Wang, Janowicz, Mai, and Majic}]{wang2024probing}
Zhangyu Wang, Krzysztof Janowicz, Gengchen Mai, and Ivan Majic. 2024.
\newblock \href {https://doi.org/10.4230/LIPIcs.COSIT.2024.9} {{Probing the
  Information Theoretical Roots of Spatial Dependence Measures}}.
\newblock In \emph{16th International Conference on Spatial Information Theory
  (COSIT 2024)}, volume 315 of \emph{Leibniz International Proceedings in
  Informatics (LIPIcs)}, pages 9:1--9:18, Dagstuhl, Germany. Schloss Dagstuhl
  -- Leibniz-Zentrum f{\"u}r Informatik.

\bibitem[{Wang et~al.(2025{\natexlab{b}})Wang, Wu, Cao, Xia, Liu, Xie, Nambi,
  Ganu, Lao, Liu, and Mai}]{wang2025geobs}
Zhangyu Wang, Nemin Wu, Qian Cao, Jiangnan Xia, Zeping Liu, Yiqun Xie,
  Akshay~Uttama Nambi, Tanuja Ganu, Ni~Lao, Ninghao Liu, and Gengchen Mai.
  2025{\natexlab{b}}.
\newblock \href {https://doi.org/10.48550/ARXIV.2509.23482} {{GeoBS}:
  Information-theoretic quantification of geographic bias in {AI} models}.
\newblock \emph{arXiv preprint arXiv:2509.23482}.

\bibitem[{{WorldPop}(2018)}]{worldpop2018}
{WorldPop}. 2018.
\newblock \href {https://doi.org/10.5258/SOTON/WP00647} {Global high resolution
  population denominators project}.
\newblock University of Southampton.

\bibitem[{Wu et~al.(2024)Wu, Cao, Wang, Liu, Qi, Zhang, Ni, Yao, Ma, Mu, Ermon,
  Ganu, Nambi, Lao, and Mai}]{wu2024torchspatial}
Nemin Wu, Qian Cao, Zhangyu Wang, Zeping Liu, Yanlin Qi, Jielu Zhang, Joshua
  Ni, Xiaobai Yao, Hongxu Ma, Lan Mu, Stefano Ermon, Tanuja Ganu, Akshay Nambi,
  Ni~Lao, and Gengchen Mai. 2024.
\newblock \href {https://doi.org/10.52202/079017-2588} {{TorchSpatial}: A
  location encoding framework and benchmark for spatial representation
  learning}.
\newblock In \emph{Advances in Neural Information Processing Systems 37}, pages
  81437--81460.

\bibitem[{Zhao et~al.(2021)Zhao, Wallace, Feng, Klein, and
  Singh}]{zhao2021calibrate}
Zihao Zhao, Eric Wallace, Shi Feng, Dan Klein, and Sameer Singh. 2021.
\newblock \href {https://proceedings.mlr.press/v139/zhao21c.html} {Calibrate
  before use: Improving few-shot performance of language models}.
\newblock In \emph{Proceedings of the 38th International Conference on Machine
  Learning}, volume 139 of \emph{Proceedings of Machine Learning Research},
  pages 12697--12706. PMLR.

\bibitem[{Zheng et~al.(2024)Zheng, Zhou, Meng, Zhou, and
  Huang}]{zheng2024large}
Chujie Zheng, Hao Zhou, Fandong Meng, Jie Zhou, and Minlie Huang. 2024.
\newblock \href {https://openreview.net/forum?id=shr9PXz7T0} {Large language
  models are not robust multiple choice selectors}.
\newblock In \emph{The Twelfth International Conference on Learning
  Representations}.

\bibitem[{Zheng et~al.(2023)Zheng, Chiang, Sheng, Zhuang, Wu, Zhuang, Lin, Li,
  Li, Xing, Zhang, Gonzalez, and Stoica}]{zheng2023judging}
Lianmin Zheng, Wei-Lin Chiang, Ying Sheng, Siyuan Zhuang, Zhanghao Wu, Yonghao
  Zhuang, Zi~Lin, Zhuohan Li, Dacheng Li, Eric Xing, Hao Zhang, Joseph
  Gonzalez, and Ion Stoica. 2023.
\newblock \href {https://doi.org/10.52202/075280-2020} {Judging
  {LLM-as-a-Judge} with {MT-Bench} and {Chatbot Arena}}.
\newblock In \emph{Advances in Neural Information Processing Systems 36}, pages
  46595--46623.

\end{thebibliography}

\appendix
\section{Data Contamination}
\label{app:contamination}

Both source benchmarks in \dataset{}, LocBench \citep{wu2024torchspatial} and the location-rating benchmark of \citet{manvi2024large}, were publicly released prior to all evaluated models, and evaluation queries sent to closed-source APIs may be retained for subsequent training \citep{balloccu2024leak}.
Observed accuracy or apparent fairness could therefore reflect two distinct confounders rather than genuine spatial awareness or unbiased behavior \citep{sainz2023nlp}: \emph{verbatim item memorization}, which inflates task performance, or \emph{benchmark-specific tuning}, where models are aligned to appear unbiased specifically under the published prompt template.
Because none of the evaluated models disclose their training corpora, we conduct three complementary black-box diagnostics, each contrasting the published benchmark against a matched control designed to reduce the risk of memorization: \one{} option-order sensitivity on Task~1, \two{} a data contamination quiz on Task~2, and \three{} template rewording on Task~2.
Across all three tests, we find minimal evidence of benchmark memorization or template-specific leakage, apart from a small original-order gain for \mluna{}.

\begin{table}[t]
  \centering
  \small
  \setlength{\tabcolsep}{3pt}
  \resizebox{\columnwidth}{!}{%
    \begin{tabular}{@{}lcccc@{}}
      \toprule
      \textbf{Model}            & \textbf{SD} & \textbf{Orig.\ $-$ Shuf.} & \textbf{$\Delta$Orig.\ Only} & \textbf{Agree} \\
      \midrule
      \logo{qwen}\mqwenbase{}   & 0.51        & $-$0.6 to $+$0.9          & $+$0.2                       & 0.55           \\
      \logo{gemma}\mgemmabase{} & 0.41        & $-$0.7 to $+$1.3          & $+$0.5                       & 0.57           \\
      \midrule
      \logo{qwen}\mpplx{}       & 0.64        & $-$0.9 to $-$0.2          & $-$0.4                       & 0.73           \\
      \logo{qwen}\mglide{}      & 0.40        & $-$0.2 to $+$0.4          & $+$0.1                       & 0.80           \\
      \logo{gemma}\mknowline{}  & 0.53        & $-$0.2 to $+$1.2          & $+$0.5                       & 0.66           \\
      \logo{gemma}\msieve{}     & 0.31        & $-$0.3 to $-$0.1          & 0.0                          & 0.80           \\
      \midrule
      \logo{openai}\mluna{}     & 0.74        & $+$1.1 to $+$1.5          & $+$0.7                       & 0.51           \\
      \logo{typesafe}\mjev{}    & 0.52        & $-$1.8 to $+$0.8          & $-$0.4                       & 0.68           \\
      \bottomrule
    \end{tabular}}
  \caption{Option-order sensitivity in the Image and Image + Coordinates settings of Task~1. \textbf{SD}: median standard deviation of accuracy over three shuffled orders (\%). \textbf{Orig.\ $-$ Shuf.}: original-order minus mean shuffled accuracy on fMoW and YFCC (\%). \textbf{$\Delta$Orig.\ Only}: share of samples correct only in the original order minus only under every shuffle (\%). \textbf{Agree}: share of samples with the same prediction under all shuffles.}
  \label{tab:contam_order}
\end{table}

\subsection{Option-Order Sensitivity}
\label{app:contam_order}

Models that have memorized multiple-choice items tend to rely on the canonical ordering of candidate labels seen during training, resulting in measurable accuracy drops when options are permuted \citep{song2024mmdetect,zheng2024large}.
To test for this effect on Task~1, we re-evaluate all eight models in the Image and Image + Coordinates settings on the same 2,000-sample subsets used for the closed-source models.
Each subset is evaluated under three random label permutations and, for fMoW and YFCC, under the original unshuffled dataset class order.
We exclude iNat2017 and iNat2018 from the original-order comparison because their candidate shortlists are ranked by the TorchSpatial image classifier's confidence, which would directly leak the ground-truth label if left unshuffled.

Table~\ref{tab:contam_order} shows no signs of item memorization for the open-weight models.
Across all 24 of their model--dataset--setting comparisons on fMoW and YFCC, original-order top-1 accuracy differs from the shuffled mean by $-$0.9 to $+$1.3\% (overall mean $+$0.05\%, higher in only 8 of 24 cases), falling within the natural variation across random permutations (median standard deviation 0.44\%).
Item-level error patterns strengthen this finding: the proportion of samples answered correctly \emph{only} under the original order closely matches the proportion answered correctly \emph{only} across all three shuffled orders (net difference between $-$0.4\% and $+$0.5\%).
Beyond ruling out memorization, these results indicate that decision fine-tuning substantially reduces permutation sensitivity: the open-weight decision models maintain consistent predictions on 66--80\% of samples, whereas the base LLMs output the same prediction across all three random orderings on 55--57\% of samples.
\mluna{} is the exception: its original-order accuracy exceeds the shuffled mean in all four comparisons ($+$1.1 to $+$1.5\%, about 1.4--2.8 standard deviations), and samples correct only in the original order are slightly more frequent ($+$0.7\%).
This small but consistent gain may reflect exposure to the original class order, or an interaction between the fixed order and \mluna{}'s strong position bias (Appendix~\ref{app:contam_quiz}). Our tests cannot distinguish the two.
Conversely, \mjev{} shows no such gain: its original-order accuracy is $-$1.8 to $+$0.8\% from the shuffled mean, lower in three of four comparisons, with fewer samples correct only in the original order ($-$0.4\%).

\begin{table}[t]
  \centering
  \small
  \setlength{\tabcolsep}{3pt}
  \begin{tabular}{@{}llccc@{}}
    \toprule
    \textbf{Model}                          & \textbf{Source} & \textbf{Picked} & \textbf{$P$(orig.)} & \textbf{Wins} \\
    \midrule
    \multirow{2}{*}{\logo{openai}\mluna{}}  & Published       & 26.1            & 0.278               & 47.2          \\
                                            & Control         & 26.0            & 0.276               & 47.2          \\
    \midrule
    \multirow{2}{*}{\logo{typesafe}\mjev{}} & Published       & 28.6            & 0.266               & 38.5          \\
                                            & Control         & 30.6            & 0.266               & 31.0          \\
    \bottomrule
  \end{tabular}
  \caption{Contamination quiz on 200 Task~2 locations. \textbf{Published}: released descriptions; \textbf{Control}: never-published descriptions. \textbf{Picked}: share of requests choosing the original (\%). \textbf{$P$(orig.)}: mean probability of the original. \textbf{Wins}: share of items where the original is most probable after averaging over positions (\%).}
  \label{tab:contam_quiz}
\end{table}

\subsection{Data Contamination Quiz}
\label{app:contam_quiz}

To test whether the closed-source models (\mluna{} and \mjev{}) memorized Task~2 location descriptions verbatim, we adopt the Data Contamination Quiz \citep{golchin2023quiz}, asking each model to identify which of four candidate texts ``appears word for word in a publicly released dataset''.
For 200 Task~2 locations, the four options comprise the published location description (coordinates, address, and nearby places) alongside three near-duplicate distractors constructed by perturbing the fifth decimal digit of the coordinates and shifting nearby-place distances by 0.1\,km while preserving distance order.
Because models may prefer unperturbed text for stylistic or formatting reasons rather than memorization, we construct a matched \emph{control} quiz from 200 never-published composite descriptions, pairing each location's coordinates and address with the nearby places of another location, and apply the same perturbations.
To control for option-position bias, which is severe in \mluna{} (selecting the first option in 97\% of queries, versus 51\% for \mjev{}), each item is queried with the unperturbed text placed in every one of the four positions, and option probabilities are averaged across rotations.

As reported in Table~\ref{tab:contam_quiz}, both models select the unperturbed text above the 25\% random-chance baseline, but do so at virtually identical rates on the never-published control.
After position averaging, \mluna{} assigns the highest probability to the unperturbed option on 47.2\% of published items versus 47.2\% of control items (mean probability 0.278 vs.\ 0.276).
Likewise, \mjev{} favors the unperturbed option on 38.5\% of published items versus 31.0\% of control items, a difference that is not statistically significant (exact McNemar test, $p=0.16$), with identical mean probabilities (0.266 vs.\ 0.266) and a slightly higher raw pick rate on the control (30.6\% vs.\ 28.6\%).
The above-chance preference therefore reflects surface-form cues shared by the control rather than verbatim memorization.
We also tested the published prompt preamble of \citet{manvi2024large} against three meaning-preserving rewordings. Both models favor the original preamble (mean probability 0.48 for \mluna{} and 0.72 for \mjev{}), though as a single item without a matched control this preference is inconclusive and may reflect fluency.

\begin{table}[t]
  \centering
  \small
  \setlength{\tabcolsep}{3pt}
  \resizebox{\columnwidth}{!}{%
    \begin{tabular}{@{}l *{6}{>{\centering\arraybackslash}p{3em}}@{}}
      \toprule
                         & \multicolumn{2}{c}{$\rho$} & \multicolumn{2}{c}{\textbf{MAD}} & \multicolumn{2}{c@{}}{\textbf{Answer rate}}                                        \\
      \cmidrule(lr){2-3}\cmidrule(lr){4-5}\cmidrule(l){6-7}
      \textbf{Template}  & \textbf{C}                 & \textbf{P}                       & \textbf{C}                                  & \textbf{P} & \textbf{C} & \textbf{P} \\
      \midrule
      \multicolumn{7}{@{}l}{\raisebox{-0.2em}[0.75em][0pt]{\includegraphics[height=0.95em]{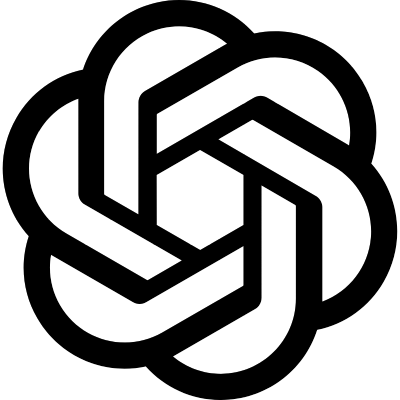}}\hspace{0.35em}\mluna{}}                                             \\
      Original           & $-$0.48                    & $-$0.62                          & 0.10                                        & 0.10       & 0.90       & 0.88       \\
      Reworded preamble  & $-$0.41                    & $-$0.42                          & 0.07                                        & 0.06       & 0.96       & 0.79       \\
      Paraphrased topics & $-$0.58                    & $-$0.60                          & 0.12                                        & 0.10       & 1.00       & 0.92       \\
      1--5 scale         & $-$0.21                    & $-$0.31                          & 0.27                                        & 0.30       & 0.83       & 0.83       \\
      Reversed scale     & $-$0.49                    & $-$0.01                          & 0.14                                        & 0.12       & 0.39       & 0.64       \\
      \midrule
      \multicolumn{7}{@{}l}{\raisebox{-0.2em}[0.75em][0pt]{\includegraphics[height=0.95em]{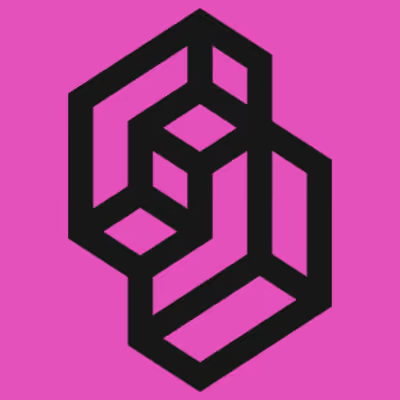}}\hspace{0.35em}\mjev{}}                                            \\
      Original           & $-$0.75                    & $-$0.90                          & 0.59                                        & 0.70       & 1.00       & 1.00       \\
      Reworded preamble  & $-$0.75                    & $-$0.89                          & 0.54                                        & 0.74       & 1.00       & 1.00       \\
      Paraphrased topics & $-$0.73                    & $-$0.83                          & 0.51                                        & 0.41       & 1.00       & 1.00       \\
      1--5 scale         & $-$0.72                    & $-$0.86                          & 0.76                                        & 0.95       & 1.00       & 1.00       \\
      Reversed scale     & $-$0.76                    & $-$0.87                          & 0.41                                        & 0.33       & 1.00       & 1.00       \\
      \bottomrule
    \end{tabular}}
  \caption{Template rewording on Task~2 (500 locations; intelligence) in the Coordinates (C) and Place Name (P) settings. $\rho$: Spearman correlation of ratings with infant mortality. \textbf{MAD}: mean absolute deviation of ratings. The other subjective topics follow the same pattern.}
  \label{tab:contam_reword}
\end{table}

\subsection{Template Rewording}
\label{app:contam_reword}

Template-based bias evaluations can be brittle to meaning-preserving edits \citep{seshadri2022quantifying}. A model aligned specifically against a published benchmark would appear unbiased only under its exact template wording.
This diagnostic is particularly pertinent for \mluna{}, whose near-zero social bias score $B$ in Section~\ref{sec:rq2} stems from tightly compressed ratings around 5.0.
We re-evaluate both closed-source models on a fixed 500-location subset of Task~2 in the Coordinates and Place Name settings across all five subjective topics and two objective controls (average temperature and infant mortality).
Alongside the original template, we test four meaning-preserving variants: \one{} a reworded preamble, \two{} paraphrased topic descriptions (e.g., ``How smart the people who live here are''), \three{} a coarser 1--5 ordinal scale, and \four{} a reversed 0.0--9.9 scale where 0.0 denotes the highest level.
Ratings are mapped back to the 0.0--9.9 scale for scoring.

Table~\ref{tab:contam_reword} shows that neither model exhibits weaker bias specifically under the original template.
For \mjev{}, intelligence ratings remain strongly and negatively correlated with infant mortality across every variant ($\rho$ between $-$0.72 and $-$0.76 with Coordinates, and between $-$0.83 and $-$0.90 with Place Name).
For \mluna{}, intelligence ratings stay negatively correlated with infant mortality under every variant except the reversed scale with place names ($\rho=-0.01$), and its rating compression on intelligence, morality, and attractiveness persists across templates ($\mathrm{MAD} \le 0.14$ on all 0.0--9.9 variants. On the 1--5 scale, $\mathrm{MAD} \le 0.30$ reflects quantization, as one scale step spans roughly 2.5 points after rescaling).
Rather than revealing template-specific tuning, rewording primarily shifts \mluna{}'s refusal rate; for instance, increasing refusals on intelligence in the Coordinates setting from 10\% under the original template to 61\% under the reversed scale.
Across variants, both models maintain strong correlations with ground truth on the objective controls (temperature $\rho \ge 0.74$; infant mortality $\rho \ge 0.54$), confirming that the perturbed prompts continue to elicit valid ratings.
The sole exception is \mjev{} on infant mortality under the reversed scale ($\rho = -0.17$ and $-0.10$), where it spreads probability mass nearly uniformly instead of inverting the scale.

\begin{figure}[t]
  \centering
  \scriptsize
  \tcbset{exbox/.style={colback=gray!4,colframe=gray!60,boxrule=0.4pt,left=3pt,right=3pt,lefttitle=2pt,righttitle=1pt,top=2pt,bottom=2pt,
        fonttitle=\scriptsize,colbacktitle=gray!25,coltitle=black,before upper={\raggedright}}}
  \begin{tcolorbox}[exbox,title={\textbf{Task 1}: Input Settings}]
    \textbf{Image:} \texttt{<image>} \\
    \textbf{Coordinates:} \texttt{Location: latitude <lat>, longitude <lon>} \\
    \textbf{Image + Coordinates:} \texttt{<image>} + \texttt{Location: latitude <lat>, longitude <lon>}
  \end{tcolorbox}
  \vspace{-4pt}
  \begin{tcolorbox}[exbox,title={\textbf{Task 1}: fMoW}]
    \textbf{Image question:} Which land use or building category does this satellite image show? \\
    \textbf{Coordinates question:} Which land use or building category is most likely to be found at this location? \\
    \textbf{Labels:} \texttt{<62 land-use and building categories>}
  \end{tcolorbox}
  \vspace{-4pt}
  \begin{tcolorbox}[exbox,title={\textbf{Task 1}: YFCC}]
    \textbf{Image question:} Which of these concepts best describes this photo? \\
    \textbf{Coordinates question:} Which of these concepts best describes a photo taken at this location? \\
    \textbf{Labels:} \texttt{<100 everyday concepts>}
  \end{tcolorbox}
  \vspace{-4pt}
  \begin{tcolorbox}[exbox,title={\textbf{Task 1}: iNat2017 / iNat2018}]
    \textbf{Image question:} Which species is shown in this photo? \\
    \textbf{Coordinates question:} Which species is most likely to be observed at this location? \\
    \textbf{Labels:} \texttt{<20 shortlisted species>}
  \end{tcolorbox}
  \vspace{-4pt}
  \begin{tcolorbox}[exbox,title={\textbf{Task 2}: Input Settings}]
    \textbf{Coordinates:} \texttt{Coordinates: (<lat>, <lon>)} \\
    \textbf{Place Name:} \texttt{Address: "<address>"} \\
    \textbf{Full:} \newline \texttt{Coordinates: (<lat>, <lon>)} \newline \texttt{Address: "<address>"} \newline \texttt{Nearby Places: "<places>"}
  \end{tcolorbox}
  \vspace{-4pt}
  \begin{tcolorbox}[exbox,title={\textbf{Task 2}: Question and Labels}]
    \textbf{Question:} You will be given data about a specific location randomly sampled from all human-populated locations on Earth. You give your rating keeping in mind that it is relative to all other human-populated locations on Earth (from all continents, countries, etc.). Rate: \texttt{<topic>} (on a scale from 0.0 to 9.9). \\
    \textbf{Labels:} 0.0, 0.1, \ldots, 9.9
  \end{tcolorbox}
  \caption{Prompt templates of \dataset{}. In Task~1, the Image and Image + Coordinates settings use the image question, and the Coordinates setting uses the coordinates question. In Task~2, all input settings and topics share one question, adapted from \citet{manvi2024large} without its output-format instruction.}
  \label{fig:prompts}
\end{figure}

\subsection{Summary and Scope}
\label{app:contam_summary}

Overall, our three black-box diagnostics show little evidence of data leakage: original and shuffled label orders yield equivalent Task~1 accuracy for the open-weight models, with only a small original-order gain for \mluna{}, published and never-published Task~2 descriptions are favored equally often, and the bias profiles of both closed-source models persist across almost all template perturbations.
Nevertheless, black-box testing cannot definitively rule out every form of indirect pretraining exposure. Moreover, the contamination quiz and template rewording cover only Task~2 and the closed-source models, motivated by the fact that \mluna{} is the only model with a near-zero bias score and \mjev{} is also closed-sourced.

\section{Prompt Templates}
\label{app:prompts}

Figure~\ref{fig:prompts} shows the prompt templates of Task~1 and Task~2.

\section{Task 2 Ground Truth}
\label{app:groundtruth}

\begin{table*}[t]
\centering
\small
\setlength{\tabcolsep}{2.6pt}
\resizebox{\textwidth}{!}{%
\begin{tabular}{@{}l *{4}{>{\centering\arraybackslash}p{2.5em}}|*{4}{>{\centering\arraybackslash}p{2.5em}}|*{4}{>{\centering\arraybackslash}p{2.5em}}|*{4}{>{\centering\arraybackslash}p{2.5em}}@{}}
\toprule
 & \multicolumn{4}{c|}{\textbf{YFCC}} & \multicolumn{4}{c|}{\textbf{fMoW}} & \multicolumn{4}{c|}{\textbf{iNat2017}} & \multicolumn{4}{c@{}}{\textbf{iNat2018}} \\
\cmidrule(lr){2-5}\cmidrule(lr){6-9}\cmidrule(lr){10-13}\cmidrule(lr){14-17}
\textbf{Model} & \textbf{I} & \textbf{C} & \textbf{I+C} & \textbf{I$\times$P} & \textbf{I} & \textbf{C} & \textbf{I+C} & \textbf{I$\times$P} & \textbf{I} & \textbf{C} & \textbf{I+C} & \textbf{I$\times$P} & \textbf{I} & \textbf{C} & \textbf{I+C} & \textbf{I$\times$P} \\
\midrule
TorchSpatial & \underline{5.9} & 9.4 & N/A$^*$ & \textbf{5.2} & \underline{333} & 202 & N/A$^*$ & \textbf{331} & 40.1 & 96.6 & N/A$^*$ & 53.9 & 23.3 & 11.2 & N/A$^*$ & 28.4 \\
\midrule
\logo{qwen}\mqwenbase{} & 15.4 & 15.1 & 18.6 & 16.4 & 336 & 2.0 & 338 & 344 & 20.5 & 24.7 & 15.1 & \underline{12.1} & 37.2 & 45.3 & 35.1 & \textbf{4.9} \\
\logo{gemma}\mgemmabase{} & 17.1 & 6.1 & 18.6 & 16.6 & 345 & 11.0 & 339 & 343 & 28.5 & 27.3 & 19.9 & \textbf{9.1} & 48.9 & 34.4 & 47.0 & 8.7 \\
\midrule
\logo{qwen}\mpplx{} & 19.5 & 15.3 & 25.4 & 22.9 & 363 & 8.5 & 365 & 365 & 16.2 & 71.4 & 12.5 & 16.3 & 34.2 & 69.3 & 30.3 & 6.4 \\
\logo{qwen}\mglide{} & 26.6 & 23.2 & 27.6 & 26.3 & 363 & 22.3 & 363 & 365 & 20.2 & 93.2 & 13.4 & 17.2 & 35.8 & 75.0 & 30.6 & 7.2 \\
\logo{gemma}\mknowline{} & 21.9 & 9.0 & 17.7 & 21.6 & 354 & 24.5 & 354 & 354 & 30.3 & 35.9 & 18.9 & 12.5 & 53.9 & 43.5 & 54.3 & 7.3 \\
\logo{gemma}\msieve{} & 28.2 & 12.5 & 29.0 & 26.6 & 370 & 12.3 & 373 & 373 & 23.5 & 23.1 & 15.3 & 16.5 & 47.1 & 34.4 & 46.5 & \underline{6.3} \\
\bottomrule
\end{tabular}}
\caption{Shuffle-corrected SSI scores, \ssiex{}, ($\downarrow$) on Task~1 in the Image (I), Coordinates (C), Image + Coordinates (I+C), and Image $\times$ Location Prior (I$\times$P) settings. Bold and underlined values indicate the \textbf{lowest} and \underline{second-lowest} scores per dataset, respectively; the Coordinates setting is not ranked, since near-random predictions err almost everywhere. $^*$TorchSpatial image models do not take coordinates as textual input.}
\label{tab:ssi_full}
\end{table*}

\begin{table*}[t]
\centering
\small
\setlength{\tabcolsep}{2.6pt}
\resizebox{\textwidth}{!}{%
\begin{tabular}{@{}l *{4}{>{\centering\arraybackslash}p{2.5em}}|*{4}{>{\centering\arraybackslash}p{2.5em}}|*{4}{>{\centering\arraybackslash}p{2.5em}}|*{4}{>{\centering\arraybackslash}p{2.5em}}@{}}
\toprule
 & \multicolumn{4}{c|}{\textbf{YFCC}} & \multicolumn{4}{c|}{\textbf{fMoW}} & \multicolumn{4}{c|}{\textbf{iNat2017}} & \multicolumn{4}{c@{}}{\textbf{iNat2018}} \\
\cmidrule(lr){2-5}\cmidrule(lr){6-9}\cmidrule(lr){10-13}\cmidrule(lr){14-17}
\textbf{Model} & \textbf{I} & \textbf{C} & \textbf{I+C} & \textbf{I$\times$P} & \textbf{I} & \textbf{C} & \textbf{I+C} & \textbf{I$\times$P} & \textbf{I} & \textbf{C} & \textbf{I+C} & \textbf{I$\times$P} & \textbf{I} & \textbf{C} & \textbf{I+C} & \textbf{I$\times$P} \\
\midrule
TorchSpatial & \underline{.005} & .067 & N/A$^*$ & \textbf{.003} & .109 & .634 & N/A$^*$ & .114 & .005 & .013 & N/A$^*$ & .004 & .007 & .007 & N/A$^*$ & .007 \\
\midrule
\logo{qwen}\mqwenbase{} & .007 & .061 & .006 & .009 & .084 & .019 & .079 & .076 & \textbf{.002} & .013 & .004 & .005 & .005 & .019 & .006 & \textbf{.003} \\
\logo{gemma}\mgemmabase{} & .009 & .024 & .009 & .009 & \underline{.074} & .068 & \textbf{.066} & \underline{.074} & \underline{.003} & .016 & \underline{.003} & \underline{.003} & .008 & .038 & .008 & \textbf{.003} \\
\midrule
\logo{qwen}\mpplx{} & .010 & .082 & .009 & .009 & .109 & .050 & .114 & .090 & \textbf{.002} & .007 & \underline{.003} & .005 & \textbf{.003} & .014 & .006 & \textbf{.003} \\
\logo{qwen}\mglide{} & .007 & .092 & .008 & .008 & .129 & .148 & .134 & .106 & \underline{.003} & .009 & \underline{.003} & .005 & .006 & .014 & .005 & \underline{.004} \\
\logo{gemma}\mknowline{} & .007 & .015 & .008 & .009 & .089 & .159 & .092 & .087 & \underline{.003} & .011 & \underline{.003} & .005 & .007 & .030 & .009 & .005 \\
\logo{gemma}\msieve{} & .008 & .049 & .008 & .009 & .112 & .069 & .121 & .102 & \underline{.003} & .018 & .004 & .004 & .006 & .053 & .007 & \underline{.004} \\
\bottomrule
\end{tabular}}
\caption{Shuffle-corrected SRE scores, \sreex{}, ($\downarrow$) on Task~1 in the Image (I), Coordinates (C), Image + Coordinates (I+C), and Image $\times$ Location Prior (I$\times$P) settings. Bold and underlined values indicate the \textbf{lowest} and \underline{second-lowest} scores per dataset, respectively; the Coordinates setting is not ranked, since near-random predictions err almost everywhere. $^*$TorchSpatial image models do not take coordinates as textual input.}
\label{tab:sre_full}
\end{table*}

Objective-topic ground truth is extracted from six raster datasets: infant mortality rate from CIESIN subnational v2.01 \citep{ciesin2021imr}, population density from WorldPop 2020 at 1\,km resolution \citep{worldpop2018}, built-up ratio from GHSL R2018A (2014) \citep{corbane2019ghsl}, nighttime lights from VIIRS VNL v2 (2016) \citep{elvidge2021viirs}, and average temperature (BIO1) and annual precipitation (BIO12) from CHELSA v1.2 \citep{karger2017chelsa}.
We aggregate population density by summing and all other topics by averaging.

Following GeoLLM \citep{manvi2024geollm}, we extract values from a $25\times25$-pixel window around each location, covering approximately 25\,km on 30-arcsecond rasters; for the 15-arcsecond nighttime-lights raster, we use a $49\times49$ window to cover a comparable area.
Unlike GeoLLM, which sums all non-negative pixels in each window, we mask declared nodata values and the undeclared fill codes $-9999$ and $-7777$ in the infant-mortality raster, and average rates and fractions, which would otherwise shrink near coastlines and raster edges.
We also retain negative values: CHELSA stores temperature in tenths of a degree Celsius, so GeoLLM's non-negative filter would exclude every location with a sub-zero mean temperature.
For rasters not in EPSG:4326 (GHSL), we reproject the coordinates.

We verified that our prompts and ground-truth extraction reproduce GeoLLM's, and that our bias score on GeoLLM's released GPT-3.5 attractiveness ratings ($B=0.33$ with the infant-survival anchor) is consistent with the value reported by \citet{manvi2024large}.
\citet{manvi2024large} anchor the bias score on infant survival, whereas we anchor it on infant mortality, its complement.
Since the bias score uses Spearman correlation, reversing the anchor's rank order flips the sign of $B$ without changing its magnitude.
Thus, $B<0$ in our setting corresponds to bias against locations with lower socioeconomic conditions, whereas the same bias yields $B>0$ under the infant-survival anchor.

\section{Geo-Bias Score Details}
\label{app:excess}

\paragraph{Implementation.}
We port SSI and the scale-grid SRE from the GeoBS reference implementation \citep{wang2025geobs}, whose code names SRE as SRI, and verify SSI against its saved outputs.
We make three corrections to SRE: \one{} neighborhoods index dataset samples rather than positions within each center's nearest-neighbor list; \two{} every cell's histogram uses the neighborhood's bin edges rather than bins fitted to the cell's own range; and \three{} direction sectors use the initial great-circle bearing.
We report SRE with the scale-grid partition.

\begin{table*}[t]
    \centering
    \small
    \setlength{\tabcolsep}{2.6pt}
    \resizebox{\textwidth}{!}{%
        \begin{tabular}{@{}l *{4}{>{\centering\arraybackslash}p{2.5em}}|*{4}{>{\centering\arraybackslash}p{2.5em}}|*{4}{>{\centering\arraybackslash}p{2.5em}}|*{4}{>{\centering\arraybackslash}p{2.5em}}@{}}
            \toprule
                                      & \multicolumn{4}{c|}{\textbf{YFCC}} & \multicolumn{4}{c|}{\textbf{fMoW}} & \multicolumn{4}{c|}{\textbf{iNat2017}} & \multicolumn{4}{c@{}}{\textbf{iNat2018}}                                                                                                                                                                                                                                              \\
            \cmidrule(lr){2-5}\cmidrule(lr){6-9}\cmidrule(lr){10-13}\cmidrule(lr){14-17}
            \textbf{Model}            & \textbf{I}                         & \textbf{C}                         & \textbf{I+C}                           & \textbf{I$\times$P}                      & \textbf{I}       & \textbf{C}       & \textbf{I+C}     & \textbf{I$\times$P} & \textbf{I}       & \textbf{C}       & \textbf{I+C}     & \textbf{I$\times$P} & \textbf{I}       & \textbf{C}       & \textbf{I+C}     & \textbf{I$\times$P} \\
            \midrule
            TorchSpatial              & 8.8                                & 327                                & N/A$^*$                                & 9.7                                      & 375              & 442              & N/A$^*$          & 378                 & 53.1             & 128              & N/A$^*$          & 89.1                & 30.1             & 15.0             & N/A$^*$          & 69.8                \\
            \midrule
            \logo{qwen}\mqwenbase{}   & 28.1& 389& 33.0& 29.9                                     & 355& 418& 354& 352                 & 31.9& 372& 20.5& 17.0                & 74.1& 223& 54.5& 9.3                 \\
            \logo{gemma}\mgemmabase{} & 25.9& 394& 29.6& 26.4                                     & 352& 415& 345& 349                 & 42.8& 393& 25.4& 14.5                & 90.7& 213& 67.2& 12.6                \\
            \midrule
            \logo{qwen}\mpplx{}       & 40.2                               & 381                                & 44.1                                   & 39.6                                     & 372              & 423              & 374              & 371                 & 23.3             & 324              & 17.4             & 20.8                & 63.4             & 184              & 41.0             & 10.0                \\
            \logo{qwen}\mglide{}      & 44.9                               & 385                                & 47.3                                   & 43.8                                     & 381              & 429              & 378              & 375                 & 26.9             & 305              & 17.8             & 21.6                & 66.5             & 174              & 41.2             & 11.2                \\
            \logo{gemma}\mknowline{}  & 36.5                               & 394                                & 34.5                                   & 35.7                                     & 360              & 425              & 359              & 358                 & 39.3             & 376              & 23.4             & 16.8                & 91.7             & 202              & 69.6             & 11.9                \\
            \logo{gemma}\msieve{}     & 47.5                               & 395                                & 50.1                                   & 47.5                                     & 376              & 426              & 378              & 377                 & 29.7             & 402              & 19.8             & 20.7                & 71.9             & 214              & 58.8             & 10.0                \\
            \bottomrule
        \end{tabular}}
    \caption{Raw SSI scores before shuffle correction on Task~1 in the Image (I), Coordinates (C), Image + Coordinates (I+C), and Image $\times$ Location Prior (I$\times$P) settings. $^*$TorchSpatial image models do not take coordinates as textual input.}
    \label{tab:ssi_raw}
\end{table*}

\begin{table*}[t]
    \centering
    \small
    \setlength{\tabcolsep}{2.6pt}
    \resizebox{\textwidth}{!}{%
        \begin{tabular}{@{}l *{4}{>{\centering\arraybackslash}p{2.5em}}|*{4}{>{\centering\arraybackslash}p{2.5em}}|*{4}{>{\centering\arraybackslash}p{2.5em}}|*{4}{>{\centering\arraybackslash}p{2.5em}}@{}}
            \toprule
                                      & \multicolumn{4}{c|}{\textbf{YFCC}} & \multicolumn{4}{c|}{\textbf{fMoW}} & \multicolumn{4}{c|}{\textbf{iNat2017}} & \multicolumn{4}{c@{}}{\textbf{iNat2018}}                                                                                                                                                                                                                                              \\
            \cmidrule(lr){2-5}\cmidrule(lr){6-9}\cmidrule(lr){10-13}\cmidrule(lr){14-17}
            \textbf{Model}            & \textbf{I}                         & \textbf{C}                         & \textbf{I+C}                           & \textbf{I$\times$P}                      & \textbf{I}       & \textbf{C}       & \textbf{I+C}     & \textbf{I$\times$P} & \textbf{I}       & \textbf{C}       & \textbf{I+C}     & \textbf{I$\times$P} & \textbf{I}       & \textbf{C}       & \textbf{I+C}     & \textbf{I$\times$P} \\
            \midrule
            TorchSpatial              & .011                               & .084                               & N/A$^*$                                & .010                                     & .116             & .645             & N/A$^*$          & .121                & .012             & .020             & N/A$^*$          & .011                & .016             & .016             & N/A$^*$          & .017                \\
            \midrule
            \logo{qwen}\mqwenbase{}   & .013& .107& .014& .015                                     & .092& .080& .087& .083                & .009& .026& .010& .012                & .014& .036& .015& .013                \\
            \logo{gemma}\mgemmabase{} & .016& .076& .016& .016                                     & .082& .129& .074& .082                & .010& .032& .010& .010                & .017& .060& .016& .011                \\
            \midrule
            \logo{qwen}\mpplx{}       & .017                               & .125                               & .015                                   & .016                                     & .117             & .121             & .122             & .098                & .009             & .017             & .010             & .011                & .012             & .026             & .014             & .013                \\
            \logo{qwen}\mglide{}      & .014                               & .129                               & .014                                   & .014                                     & .137             & .205             & .141             & .114                & .010             & .017             & .010             & .011                & .014             & .025             & .014             & .013                \\
            \logo{gemma}\mknowline{}  & .013                               & .070                               & .014                                   & .015                                     & .096             & .217             & .099             & .094                & .010             & .025             & .010             & .011                & .015             & .047             & .017             & .013                \\
            \logo{gemma}\msieve{}     & .015                               & .098                               & .015                                   & .016                                     & .119             & .131             & .128             & .109                & .010             & .032             & .010             & .011                & .015             & .069             & .016             & .013                \\
            \bottomrule
        \end{tabular}}
    \caption{Raw SRE scores before shuffle correction on Task~1 in the Image (I), Coordinates (C), Image + Coordinates (I+C), and Image $\times$ Location Prior (I$\times$P) settings. $^*$TorchSpatial image models do not take coordinates as textual input.}
    \label{tab:sre_raw}
\end{table*}

\paragraph{Parameters.}
For Task~1, SSI uses a 100\,km radius and a 4-nearest-neighbor weight matrix, as in TorchSpatial \citep{wu2024torchspatial}, with each neighborhood capped at its 400 nearest samples and about five background points per sample, the GeoBS default.
SRE follows the GeoBS example configuration, rounded to a 300\,km radius and 60\,km cells, with two histogram bins for the binary Hit@1 marks, and cells with fewer than 20 samples are skipped.
Both scores are computed around 2,000 evaluation centers per dataset, shared across models and drawn by stratified sampling on a $10^\circ$ grid with allocation proportional to the number of samples per cell, so they preserve the spatial distribution of the dataset.
The shuffle correction uses one fixed permutation, shared across models.

\begin{table}[!t]
    \centering
    \small
    \setlength{\tabcolsep}{2.4pt}
    \resizebox{\columnwidth}{!}{%
        \begin{tabular}{@{}l ccccc@{}}
            \toprule
            \textbf{Model}                & \textbf{Intel.}     & \textbf{Work}       & \textbf{Moral.}     & \textbf{Like.}      & \textbf{Attr.}      \\
            \midrule
            \logo{openai}\mgptfour{}      & \textbf{$-$0.01}    & $-$0.23             & \textbf{$-$0.01}    & $-$0.23             & \textbf{0.00}       \\
            \logo{openai}\mgptthreefive{} & $-$0.33             & \textbf{$-$0.13}    & $-$0.11             & $-$0.11             & $-$0.29             \\
            \logo{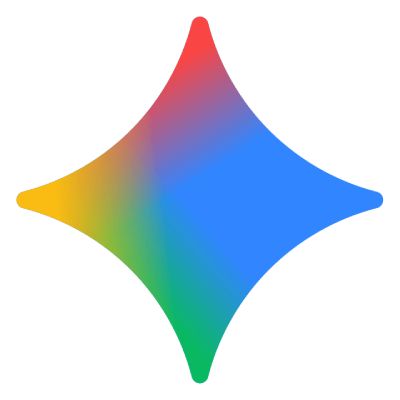}\mgeminipro{}    & $-$0.54             & $-$0.63             & $-$0.77             & $-$0.42             & $-$0.36             \\
            \logo{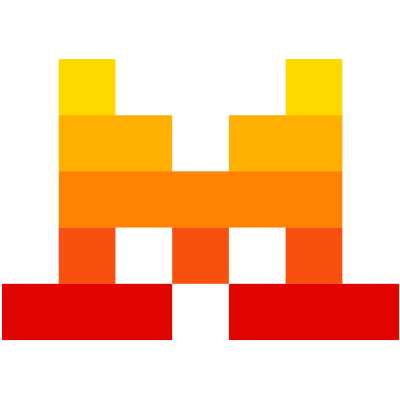}\mmixtral{}     & $-$0.19             & \underline{$-$0.16} & $-$0.38             & $-$0.43             & $-$0.32             \\
            \logo{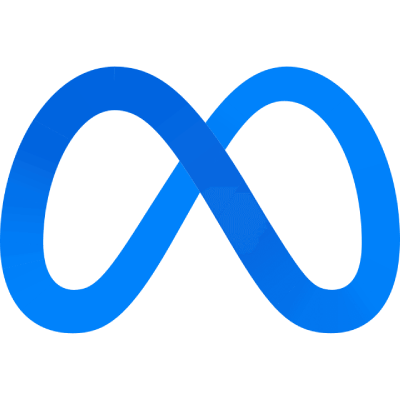}\mllama{}         & \textbf{$-$0.01}    & $-$0.23             & \underline{$-$0.02} & \textbf{$-$0.05}    & $-$0.13             \\
            \midrule
            \logo{qwen}\mqwenbase{}       & $-$0.50             & $-$0.47             & $-$0.26             & $-$0.17             & $-$0.19             \\
            \logo{gemma}\mgemmabase{}     & $-$0.73             & $-$0.48             & $-$0.09             & $-$0.08             & \underline{0.01}    \\
            \midrule
            \logo{qwen}\mpplx{}           & $-$0.51             & $-$0.37             & $-$0.32             & $-$0.17             & $-$0.11             \\
            \logo{qwen}\mglide{}          & $-$0.45             & $-$0.33             & $-$0.28             & $-$0.19             & $-$0.13             \\
            \logo{gemma}\mknowline{}      & $-$0.84             & $-$0.79             & $-$0.41             & $-$0.23             & $-$0.33             \\
            \logo{gemma}\msieve{}         & $-$0.68             & $-$0.47             & $-$0.26             & $-$0.30             & $-$0.33             \\
            \midrule
            \logo{openai}\mluna{}         & \underline{$-$0.05} & $-$0.17             & \textbf{$-$0.01}    & \underline{$-$0.06} & \underline{$-$0.01} \\
            \logo{typesafe}\mjev{}        & $-$0.64             & $-$0.28             & $-$0.18             & $-$0.13             & $-$0.19             \\
            \bottomrule
        \end{tabular}}
    \caption{Bias score $B$ ($\rightarrow 0$) on the subjective topics of Task 2 in the Full setting, compared with the generative LLMs of \citet{manvi2024large}, whose scores are sign-flipped to our infant-mortality anchor (Appendix~\ref{app:groundtruth}). Bold and underlined values indicate the \textbf{least} and \underline{second-least} biased results per topic, respectively.}
    \label{tab:manvi_baselines}
\end{table}

\paragraph{Full Results.}
Tables~\ref{tab:ssi_full} and~\ref{tab:sre_full} report the shuffle-corrected scores in all four input settings, and Tables~\ref{tab:ssi_raw} and~\ref{tab:sre_raw} the corresponding raw scores before correction.

\paragraph{Comparison with Generative LLMs.}
Table~\ref{tab:manvi_baselines} compares the bias scores of the decision models in the Full setting, which uses the same prompt format as \citet{manvi2024large}, with those that \citet{manvi2024large} report for generative LLMs.

\end{document}